\documentclass{article}

 \usepackage[main, final]{neurips_2026}

\usepackage{wrapfig}
\usepackage[utf8]{inputenc}
\usepackage[T1]{fontenc}
\usepackage{hyperref}
\usepackage{url}
\usepackage{booktabs}
\usepackage{amsfonts}
\usepackage{nicefrac}
\usepackage{microtype}
\usepackage[table]{xcolor}
\definecolor{lightblue}{RGB}{220,225,255}
\definecolor{deepblue}{RGB}{45,95,168}
\definecolor{softgray}{RGB}{90,90,90}
\definecolor{deepbrick}{RGB}{150,70,50}
\definecolor{mypurple}{RGB}{128,0,128}
\definecolor{lightpurple}{RGB}{240,235,250}
\definecolor{mygreen}{RGB}{54,146,76}
\definecolor{secondcell}{RGB}{242,251,245}
\definecolor{bestcell}{RGB}{210,238,218}
\usepackage{graphicx}
\usepackage{amsmath}
\usepackage{algorithm}
\usepackage{algorithmic}
\usepackage{marvosym}
\definecolor{commentgreen}{rgb}{0.29, 0.85, 0.49}

\usepackage{makecell}
\usepackage{caption}
\hypersetup{
    colorlinks=true,
    linkcolor=deepbrick,
    citecolor=deepblue,
    urlcolor=softgray
}

\usepackage{multirow}
\usepackage{bm}
\usepackage{siunitx}
\usepackage{pifont}
\newcommand{\gain}[1]{{\scriptsize\color{gray!85} [#1]}}
\newcommand{\ogain}[1]{\textbf{{\scriptsize\color{deepblue!100} [#1]}}}

\newcommand{\res}[2]{#1 \gain{#2}}

\newcommand{\bestres}[2]{\cellcolor{bestcell}#1 \gain{#2}}
\newcommand{\secondres}[2]{\cellcolor{secondcell}#1 \gain{#2}}

\newcommand{\obestres}[2]{\cellcolor{bestcell}#1 \ogain{#2}}
\newcommand{\osecondres}[2]{\cellcolor{secondcell}#1 \ogain{#2}}

\newcommand{\vanilla}[1]{\textcolor{gray}{#1}}
\newif\ifshowrevision \showrevisionfalse
\DeclareRobustCommand{\revcolor}{\ifshowrevision\color{red}\fi}
\definecolor{linkpink}{RGB}{255,20,147}
\DeclareRobustCommand{\new}[1]{{\revcolor #1}}
\newcommand{\appref}[1]{\textcolor{deepbrick}{Appendix}~\ref{#1}}

\title{Visual Latents Know More Than They Say: Unsilencing Latent Reasoning in MLLMs}

\author{Xin Zhang\textsuperscript{1} \quad
Qiqi Tao\textsuperscript{1,2} \quad
Jiawei Du\textsuperscript{1}\quad
Moyun Liu\textsuperscript{3} \quad
Joey Tianyi Zhou\textsuperscript{4,1 \Letter} \\
\textsuperscript{1}{\small Institute of Advanced Intelligence and Computing,}\\ {\small Agency for Science, Technology and Research (A*STAR), Singapore}\\
\textsuperscript{2}{\small Singapore University of Technology and Design, Singapore}\\
\textsuperscript{3}{\small Huazhong University of Science and Technology, China}\\
\textsuperscript{4}{\small Singapore Management University, Singapore}\\
\texttt{\small \{zhangx7, dujw, Joey\_Zhou\}@a-star.edu.sg}\\
\texttt{\small qiqi\_tao@mymail.sutd.edu.sg, lmomoy@hust.edu.cn}\\
{\small \Letter\ Corresponding author}
}

\begin{document}
\maketitle
\begin{abstract}
Continuous latent-space reasoning offers a compact alternative to textual chain-of-thought for multimodal models, enabling high-dimensional visual evidence to be integrated without explicit reasoning tokens. However, we identify a previously overlooked optimization pathology in existing latent visual reasoning methods: although visual latents become semantically enriched during training, their contribution to final answer prediction is systematically suppressed. Within the shared parameter space, the autoregressive objective favors shortcut reliance on direct visual input, driving latent tokens toward transition-like states rather than informative reasoning content. We term this phenomenon \emph{\textbf{Silenced Visual Latents}}. To address it, we disentangle the two conflicting objectives by directly optimizing the latent reasoning at inference time, keeping backbone parameters frozen. In Stage I, visual latents are warmed up via query-guided contrastive latent--visual alignment, improving semantic quality while preventing latent collapse. In Stage II, the latent reasoning is further optimized via a confidence-progression reward, which incentivizes predicted token distributions along the latent span to become progressively more concentrated, routing predictions through the latent reasoning rather than bypassing it. Experiments across eight benchmarks and four model backbones show that inference-time latent optimization, without any parameter updates, effectively unleashes the suppressed reasoning capacity of visual latents. Code is available at {\hypersetup{urlcolor=linkpink}\url{https://github.com/zhangxin-xd/Unsilencing-Latent-Reasoning}}.
\end{abstract}
\section{Introduction}
Multimodal Large Language Models (MLLMs)~\cite{yin2024survey, bai2025qwen25vltechnicalreport, chen2024internvl} reconfigure the Large Language Model (LLM) paradigm beyond text-only inputs, integrating visual encoders to enable joint reasoning over language and visual perception. While this evolution brings remarkable advances in various cross-modal reasoning tasks, such as scientific visual question answering~\cite{zou2026interns1proscientificmultimodalfoundation, weng2026deepscientist}, mathematics solving~\cite{Mini-Omni-Reasoner, gao2025omnimath}, and visual grounding~\cite{bai2025univg,tang2026visual}, the asymmetric inductive bias toward language frequently manifests as hallucinated visual understanding~\cite{wang2025imagetokens, xu2026thinkinguncertainty, wang2025mllm}. 

A straightforward remedy is to follow the standard Chain-of-Thought (CoT)~\cite{wei2022chain, wang2025multimodal} paradigm, encouraging explicit textual reasoning prior to the final answer prediction~\cite{zhang2024multimodalchainofthoughtreasoninglanguage, zhao2025promptcot, zhang2025mm}. Subsequent efforts further expand the evidential space available during reasoning by enabling active visual manipulations~\cite{shao2024visual, hu2024visual, zhou2024image} or intermediate generation~\cite{ye2024diffusion, wu2024mind, fu2025refocus}. In short, the key idea is to embed additional auxiliary reasoning contexts to maximize the probability of the correct answer. 
Another promising approach is \textit{Latent Reasoning}~\cite{map2025latent}, recently explored in LLMs (e.g., Coconut~\cite{haotraining}, CoLaR~\cite{tan2025colar}), which shifts intermediate reasoning from discrete tokens to continuous latent representations~\cite{shen2025codi,shi2026swireasoning}. By operating in latent space, this paradigm increases representational bandwidth while reducing autoregressive context length. This expanded capacity is particularly advantageous for multimodal reasoning, enabling compact integration of high-dimensional visual evidence without external tool invocation or complex pixel-level generation~\cite{li2025lvr,ma2025latentrefocusing}.

\begin{wrapfigure}{r}{0.52\linewidth}
\vspace{-1em}
\small
    \centering
    \includegraphics[width=1\linewidth]{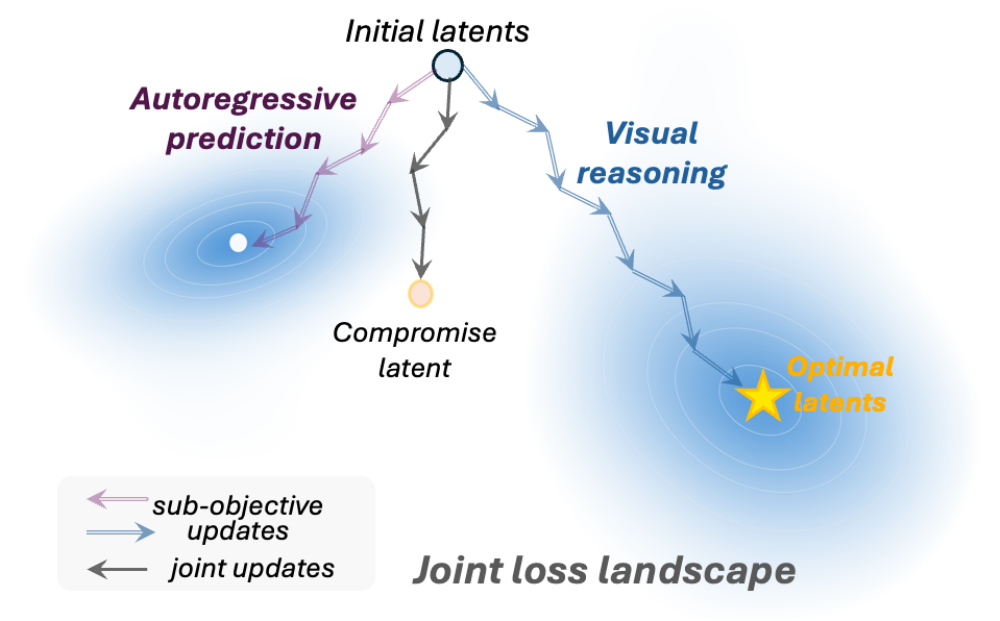}
\caption{\small The joint loss landscape of Latent Visual Reasoning. Under joint optimization, the initial latents are simultaneously pulled toward two conflicting attractors: the autoregressive prediction objective (left) and the visual reasoning objective (right). The shortcut favored by the autoregressive prediction objective dissociates latent quality from latent effectiveness and bypasses meaningful latent reasoning, driving the latents toward a compromise state.}
    \label{fig:teaser}
    \vspace{-1.2em}
\end{wrapfigure}
Current \textit{Latent Visual Reasoning (LVR)} methods instantiate latent representations by aligning them with diverse forms of pre-curated visual evidence, such as trajectory maps in Mirage~\cite{yang2025machine}, regions of interest (ROIs) in LVR~\cite{li2025lvr}, 3D geometric priors in 3DThinker~\cite{chen2025think}, and combinations of segmentation, depth, and edge maps in CoVT~\cite{qin2025chain}. Beyond mere alignment with these pre-curated informative visual targets, the latents are simultaneously optimized to ensure accurate autoregressive answer prediction.
Despite substantial improvements in multimodal reasoning, a critical inconsistency actually persists under joint optimization. Through preliminary analysis, we observe that visual latents become progressively better aligned with task-relevant visual evidence, yet answer accuracy does not improve accordingly and often fluctuates. This reveals a counterintuitive dissociation between \emph{latent quality} and \emph{latent effectiveness}: stronger latent representations are not necessarily translated into stronger final prediction behavior. As illustrated in \autoref{fig:teaser}, this is because, within the shared parameter space, the autoregressive answer prediction objective tends to favor shortcut routes through direct visual input, while latent tokens are gradually pushed toward transition-like behavior rather than semantically informative reasoning states. Thus, the resulting latent representations are learned in a compromised manner and systematically under-utilized at answer time, which we term \textbf{\emph{Silenced Visual Latents}}.

Based on this insight, we disentangle these two objectives by directly optimizing the latent reasoning at inference time, keeping backbone parameters frozen. In Stage I, we perform visual latent warm-up via query-guided contrastive latent--visual alignment: visual tokens are ranked by their relevance to the query and partitioned into chunk-wise positive and negative sets, pulling each latent toward query-relevant visual evidence while pushing it away from irrelevant patches, with chunk-wise assignment preventing latent collapse by ensuring each latent token attends to a distinct subset of visual evidence. In Stage II, we enforce latent utilization via a confidence-progression reward: the latent reasoning is iteratively perturbed and updated so that predicted token distributions along the latent span become progressively more concentrated, directly incentivizing the model to route predictions through the latent reasoning rather than bypassing it. Together, they directly address both failure modes of Silenced Visual Latents by optimizing the latent reasoning itself, without modifying any model parameters.

Our contributions are threefold: \textbf{(1)} We identify the Silenced Visual Latents phenomenon, revealing that visual latents become enriched yet remain under-utilized due to autoregressive shortcuts in the shared parameter space. \textbf{(2)} We propose an inference-time framework that disentangles latent quality from latent utilization via contrastive alignment and reward-driven reinforcement, with frozen backbone. \textbf{(3)} Extensive experiments across eight benchmarks and four model backbones confirm the effectiveness and generalizability of our framework, requiring no parameter updates.
\vspace{-1em}
\section{Related Works}
\vspace{-1em}
Latent space is broadly acknowledged to be more expressive than discrete token space, thanks to its continuous nature, inspiring a growing line of work that moves intermediate reasoning into the continuous domain. Latent reasoning in LLMs was pioneered by Coconut~\cite{haotraining}, which feeds the model's last hidden state back as input without decoding, enabling implicit breadth-first search in latent space. Subsequent works improve upon this: CODI~\cite{shen2025codi} aligns latent states with an explicit teacher via self-distillation, CoLaR~\cite{tan2025colar} dynamically compresses reasoning chains to reduce inference cost, and Deliberation~\cite{liudeliberation} augments the KV cache with differentiable latent vectors to enable iterative refinement of reasoning states without generating explicit tokens. These works collectively show that latent reasoning can be more expressive and efficient than textual CoT. 

Extending latent reasoning to multimodal settings also brings additional advantages: the model must integrate high-dimensional visual evidence into the reasoning process without resorting to costly pixel-level generation or external tool invocations. Several paradigms have recently emerged to explore this. Mirage~\cite{yang2025machine} inserts compact latent visual tokens into the autoregressive reasoning and aligns them with trajectory-map visual signals, enabling visual imagination without decoding actual images. LVR~\cite{li2025lvr} supervises latent tokens with pre-curated regions of interest (ROIs) and jointly optimizes them with answer prediction. CoVT~\cite{qin2025chain} enriches this supervision by incorporating diverse visual targets, such as segmentation, depth, and edge maps, providing richer visual grounding for latent tokens.
Furthermore, Monet~\cite{wang2025monet} improves supervision for latent visual reasoning by training MLLMs to generate continuous visual-thought embeddings through distillation-based SFT and visual-latent policy optimization.
More recently, 3DThinker~\cite{chen2025think} incorporates 3D geometric priors as visual supervision, broadening the scope of latent visual reasoning to spatial understanding tasks. Despite their diversity in supervision design, most existing methods share a common optimization scheme: latent representations are optimized jointly with autoregressive answer prediction under a shared parameter space. This coupling implicitly assumes that improving the semantic quality of visual latents will naturally translate into stronger answer prediction. However, as we show in this work, the autoregressive objective can instead favor shortcut reliance on direct visual inputs, driving visual latents toward suboptimal compromise states and causing them to be under-utilized during final prediction. We refer to this joint failure as Silenced Visual Latents.
\begin{figure}
    \centering
    \small
    \includegraphics[width=1\linewidth]{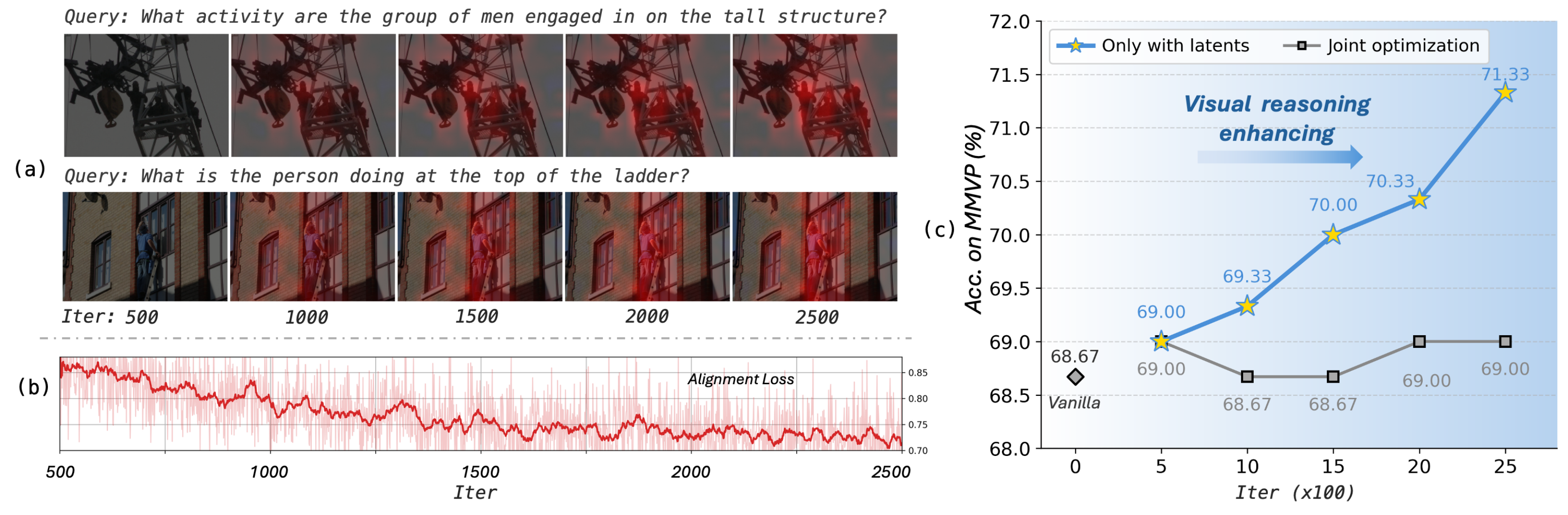}
    \vspace{-1.6em}
    \caption{\small Enhanced visual latents improve reasoning performance, but this benefit is suppressed under joint optimization. Left: Visual latents are progressively enhanced during training, as evidenced by the deepening red color indicating higher similarity between latents and pre-curated visual clues (a), while the alignment loss decreases steadily (b). Right: Although visual latents continue to improve, the jointly optimized model's performance still fluctuates (\textcolor{gray}{gray line}). In contrast, equipping the vanilla model with latents from each checkpoint yields a monotonically increasing trend (\textcolor{deepblue}{blue line}), revealing that enhanced latents do benefit reasoning, but their contribution is suppressed under joint optimization.}
    \vspace{-2em}
    \label{fig:Visual_alignment}
\end{figure}

\vspace{-0.8em}
\section{Methodology}\label{sec:methodology}
\vspace{-0.8em}
\noindent\textbf{Problem Formulation.}
Consider an MLLM $\mathcal{M} = \mathcal{F} \circ \Phi$, where $\mathcal{F} = \{f(\cdot), g(\cdot)\}$ consists of a visual encoder $f(\cdot)$\footnote{The visual encoder includes the projector; omitted for brevity.} and a text embedding layer $g(\cdot)$. These modules encode visual and textual inputs, respectively, before they are fed into the language backbone $\Phi$ parameterized by $\bm{\theta}$. For a multimodal input $\{\texttt{Vision:}\mathcal{V} = \left(\bm{v}_1, \bm{v}_2, \ldots, \bm{v}_N\right),\texttt{Query:} \mathcal{Q}\}$, $\Phi_{\bm{\theta}}$ is prompted to first produce a $T$-step reasoning sequence $\mathcal{C} = (\bm{c}_1, \bm{c}_2, \ldots, \bm{c}_T)$ and then generate the final answer $\mathcal{A} = (\bm{a}_1, \bm{a}_2, \ldots, \bm{a}_L)$. The key insight of Latent Visual Reasoning (LVR) is to enhance multimodal reasoning by lifting intermediate reasoning from discrete textual tokens $\mathcal{C}$ to continuous visual latent representations $\mathcal{H} = (\bm{h}_1, \bm{h}_2, \ldots, \bm{h}_K)$\footnote{Special tokens \texttt{<latent\_start>} and \texttt{<latent\_end>} 
are prepended and appended to $\mathcal{H}$, respectively, to delimit the latent reasoning segment.}, where $K \ll T$. Thus, current paradigms optimize $\Phi_{\bm{\theta}}$ with a joint objective that (i) \textbf{\textit{encourages visual reasoning in the latent space}}, while (ii) \textbf{\textit{enforcing faithful answer text autoregression,}}
\begin{equation}
\bm{\theta}^{*} =
\underset{\bm{\theta}}{\arg\min}
\;
\underbrace{
\frac{1}{K} \sum_{k=1}^K
\left\|\bm{h}_k-\tilde{\bm{v}}_k\right\|_2^2
}_{\color{deepblue}\textit{\normalsize{\ding{172} visual latent reasoning}}}
\underbrace{-\lambda
\sum_{l=2}^{L} \log \Phi_{\bm{\theta}}\left(\bm{a}_{l} \mid \mathcal{V}, \mathcal{Q}, \mathcal{H}, \bm{a}_{1:l-1}\right)
}_{\color{deepbrick}\textit{\normalsize{\ding{173} answer\hspace{0.1em} autoregression}}}.
\label{eq:training_objective}
\end{equation}
Here, $\tilde{\mathcal{V}} = \left(\tilde{\bm{v}}_1, \tilde{\bm{v}}_2, \ldots, \tilde{\bm{v}}_K\right)$ denotes the visual reasoning clues (\textit{e.g.} regions of interest (ROIs) suggested by LVR~\cite{li2025lvr}) for alignment. $\lambda$ is a balancing hyperparameter between the two loss terms.
\begin{figure*}
    \centering
    \includegraphics[width=1\linewidth]{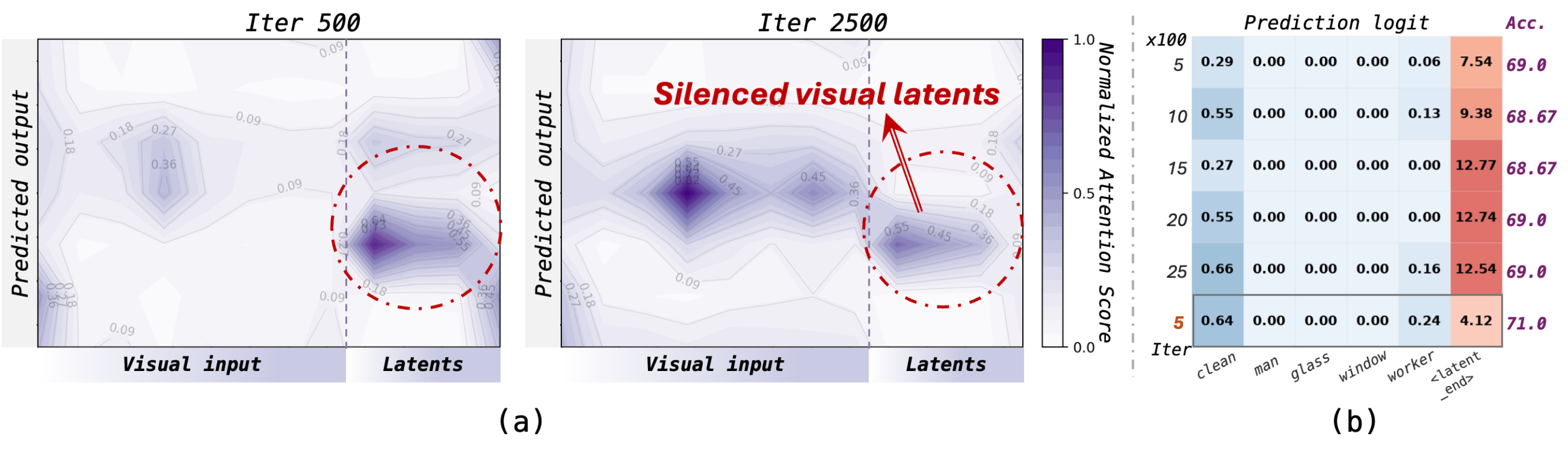}
    \vspace{-2.0em}
    \caption{\small Joint optimization silences visual latents. (a) During training, attention gradually drifts toward the input visual tokens rather than the visual latents, indicating that visual latents progressively lose their voice in answer prediction. (b) Prediction logits of the first latent token at different training stages. The latent token is increasingly pushed toward the \texttt{<latent\_end>} token, indicated by the overwhelmingly larger logit. \textcolor{orange}{5} denotes the checkpoint optimized at $500_{th}$ \textit{Iter} with half the weight of the autoregression loss, which alleviates the damage to the semantic meaning of the latent token, yielding lower \texttt{<latent\_end>} logits and higher accuracy.}
    \vspace{-2.2em}
    \label{fig:attention_logits}
\end{figure*}
\vspace{-0.8em}
\subsection{Silenced Visual Latents}\label{sec:silenced}
\vspace{-0.8em}

\textbf{Observation \#1:} \textit{\textbf{Better visual latents are the key to improving multimodal reasoning.}} \\
\noindent To understand how visual latents affect multimodal reasoning, we begin by examining what they encode. As shown in \autoref{fig:Visual_alignment} \textcolor{deepbrick}{(a)} and \textcolor{deepbrick}{(b)}, visual latents become increasingly aligned with the pre-defined visual chain as the alignment loss decreases, suggesting that they progressively capture more task-relevant visual information.
Naturally, we may ask: do such progressively enhanced visual latents foster reasoning? To isolate the effect of visual latents from the jointly optimized answer prediction branch, we use vanilla Qwen2.5VL-7B (\textit{i.e.} model at $0_{th}$ \textit{Iter}) and replace only its visual latents with those produced by later optimized checkpoints. As shown by the \textcolor{deepblue}{blue line} in \autoref{fig:Visual_alignment} \textcolor{deepbrick}{(c)}, the accuracy improves as the donated visual latents become stronger, indicating that better visual latents indeed contribute positively to multimodal reasoning.
\\\textbf{Observation \#2:} \textbf{\textit{Joint optimization silences the visual latents.}}\\
\noindent However, the benefit in Observation \#1 is not faithfully reflected in the performance of the jointly optimized checkpoints. As shown by the \textcolor{gray}{gray line} in \autoref{fig:Visual_alignment} \textcolor{deepbrick}{(c)}, the accuracy of the actual checkpoints fluctuates rather than improving monotonically over training. For example, the accuracy at the $2500_{th}$ \textit{Iter} only reaches the level of the $500_{th}$ \textit{Iter}, despite the substantially better visual latents. To further investigate this discrepancy, we visualize the attention maps of models at different training stages. As shown in \autoref{fig:attention_logits} \textcolor{deepbrick}{(a)}, after training, prediction becomes dominated by the visual input rather than the optimized latents. In addition, \autoref{fig:attention_logits} \textcolor{deepbrick}{(b)} shows that the prediction logits of latent tokens on answer-related text tokens increase during training; however, the logits on the transition token \texttt{<latent\_end>} also increase and become overwhelmingly larger. Can a smaller $\lambda$ alleviate this issue? As shown in the figure, using half of the original $\lambda$ enables the logits of answer-related tokens to reach, by the $500_{th}$ \textit{Iter}, a level comparable to that obtained at the $2500_{th}$ \textit{Iter}. Meanwhile, it yields smaller \texttt{<latent\_end>} logits, thereby improving the overall performance. These observations suggest the optimization dynamic illustrated in \autoref{fig:teaser}: both objectives are optimized within the same parameter space. As $\bm{\theta}$ is updated toward a joint optimum, the autoregressive loss rapidly exploits shortcuts grounded in the visual input, while simultaneously driving latent tokens toward a transition-token-like role at the expense of their semantic content. Consequently, joint optimization produces compromised latent tokens whose contribution is no longer effectively retained in the final answer prediction. We term this phenomenon ``\textit{\textbf{Silenced Visual Latents}}.''

As analyzed above, these two objectives are difficult to reconcile within a shared optimization space, and the resulting ``Silenced Visual Latents'' lead to unsatisfactory performance. A natural question, then, is whether their optimization spaces can be disentangled so as to preserve autoregressive answer prediction while fully exploiting the reasoning capacity of visual latents. To this end, we directly optimize the visual latents while retaining the original model parameters for autoregressive textual generation. Concretely, at test time, we perform per-instance latent-token optimization for each input sample, thereby enhancing latent effectiveness without compromising the model's original capabilities. Based on the aforementioned observations, we optimize visual latents in two stages: \textit{Stage I: Visual Latent Warm-up} and \textit{Stage II: Latent-to-Answer Reinforcement}.
\begin{figure}
\small
    \centering
    \includegraphics[width=0.96\linewidth]{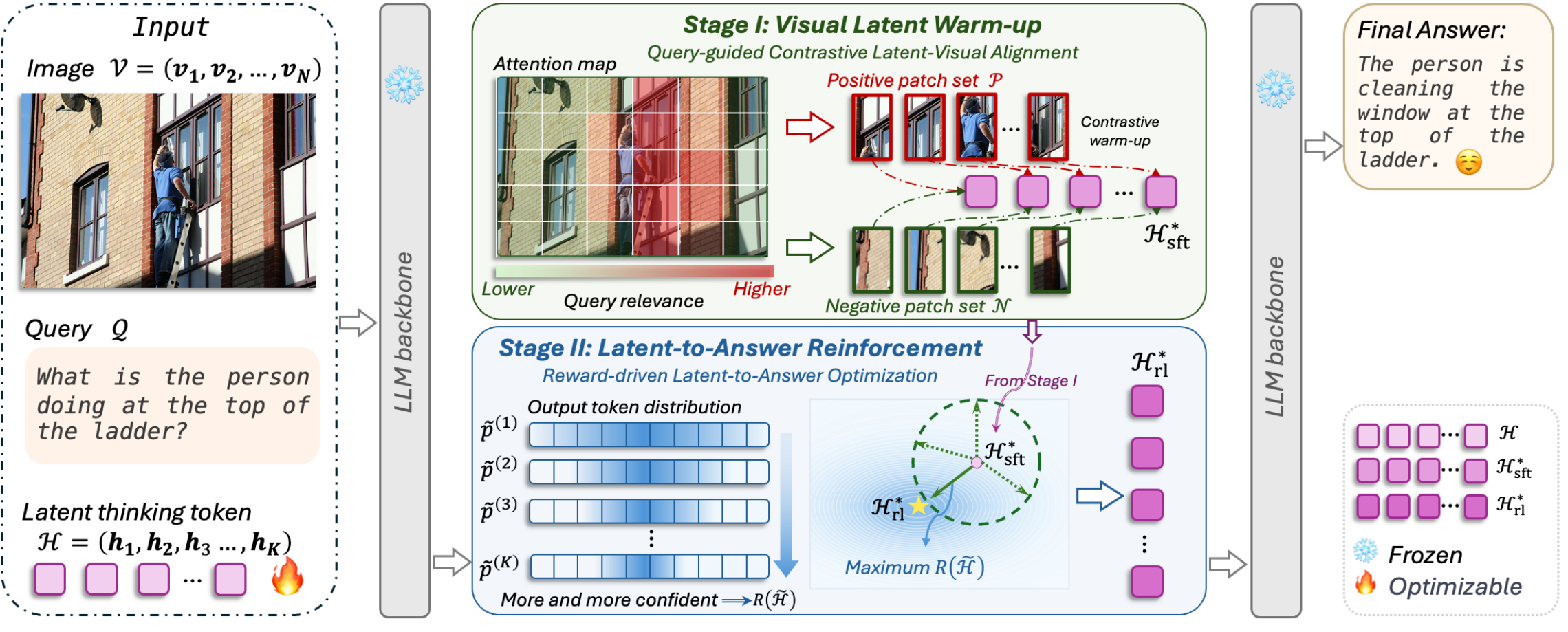}
    \vspace{-0.6em}
\caption{\small
Overview of the proposed framework for unsilencing visual latents. 
Stage I performs query-guided contrastive latent--visual alignment with positive and negative patch sets to enhance latent semantics. 
Stage II initializes from the Stage-I latents and applies a confidence-progression reward that encourages progressively more concentrated output token distributions, yielding the final latents $\mathcal{H}^{*}$ for answer generation.
}
\vspace{-1.3em}
    \label{fig:framework}
\end{figure}
\subsection{Stage I: Visual Latent Warm-up}
\label{subsubsec:stage1}
As the optimization space is disentangled, visual latents can no longer rely on the model's autoregressive generation. Accordingly, as shown in \autoref{fig:framework}, we explicitly reserve latent slots before answer generation to ensure that these tokens remain functionally involved in the reasoning process.
Stage I is designed to enhance the semantic quality of visual latents before enforcing their contribution to final answer prediction. Existing approaches~\cite{li2025lvr, wang2025monet} typically supervise latent tokens through positive alignment with visual reasoning signals. However, this positive-only supervision does not explicitly distinguish relevant visual evidence from irrelevant or misleading content. To address this limitation, we introduce a query-guided contrastive warm-up objective that imposes a discriminative constraint on each latent token. 

Let $\bm{A}\in\mathbb{R}^{M\times M}$ denote the token-level attention
matrix averaged across all layers and heads of $\Phi_{\bm{\theta}}$, where
$M$ is the total sequence length, and let
$\mathcal{I}_{\mathcal{V}}=\{1,\ldots,N\}$ and
$\mathcal{I}_{\mathcal{Q}}\subseteq\{N+1,\ldots,M\}$ denote the position
index sets of $\mathcal{V}$ and $\mathcal{Q}$ within the full sequence,
respectively.
For each visual token $\bm{v}_n \in \mathcal{V}$, its query-guided
relevance score is defined as
\begin{equation}
    s_n = \frac{1}{|\mathcal{I}_{\mathcal{Q}}|}
          \sum_{q\,\in\,\mathcal{I}_{\mathcal{Q}}} \bm{A}_{q,n},
    \qquad n \in \mathcal{I}_{\mathcal{V}},
    \label{eq:relevance_score}
\end{equation}
which measures the average attention that $\mathcal{Q}$ allocates to each
visual token in $\mathcal{V}$.

The visual tokens in $\mathcal{V}$ are ranked in descending order of $s_n$
to form a permutation $\pi$, where $\bm{v}_{\pi(1)}$ is the most
query-relevant patch with respect to $\mathcal{Q}$.
Rather than sharing a common supervision pool across all
$\bm{h}_k \in \mathcal{H}$, which would cause all latent tokens to
collapse onto the same top-ranked evidence, the ranked
tokens are partitioned into non-overlapping chunks, with distinct chunks
assigned to each $\bm{h}_k$.
Formally, the positive set $\mathcal{P}_k$ and negative set $\mathcal{N}_k$
for the $k$-th latent token $\bm{h}_k \in \mathcal{H}$ are constructed as:
\begin{equation}
    \mathcal{P}_k=\{\bm{v}_{\pi(t)}\mid t\in\mathcal{S}_k^{+}\},\qquad
    \mathcal{N}_k=\{\bm{v}_{\pi(t)}\mid t\in\mathcal{S}_k^{-}\},
    \label{eq:chunk_assignment}
\end{equation}
where $\mathcal{S}_k^{+}$ and $\mathcal{S}_k^{-}$ denote the positive and
negative index chunks assigned to the $k$-th latent token, respectively,
with all assigned chunks mutually disjoint and with fixed cardinalities
$|\mathcal{S}_k^{+}|=\texttt{pos\_num}$ and
$|\mathcal{S}_k^{-}|=\texttt{neg\_num}$ for all $k$.
This ensures that each $\bm{h}_k$ is supervised by a distinct subset of
highly relevant tokens while being repelled from a
distinct subset of low-relevance content, preserving diversity across
$\mathcal{H}$.
The latent tokens are then optimized via the following contrastive objective
over $(\mathcal{P}_k, \mathcal{N}_k)$:
\begin{equation}
    \mathcal{H}_{\mathrm{sft}}^{*}
    =
    \underset{\mathcal{H}}{\arg\min}
    \left(
        -\frac{1}{K}\sum_{k=1}^{K}
        \log
        \frac{
            \beta
            \sum_{\bm{v}\in\mathcal{P}_k}
            \exp\!\left(\mathrm{sim}(\bm{h}_k,\bm{v})/\tau\right)
        }{
            \sum_{\bm{v}\in\mathcal{P}_k\cup\mathcal{N}_k}
            \exp\!\left(\mathrm{sim}(\bm{h}_k,\bm{v})/\tau\right)
        }
    \right),
    \label{eq:stage1_loss}
\end{equation}
where $\mathrm{sim}(\cdot,\cdot)$ denotes cosine similarity, $\tau$ is
a temperature hyperparameter, and
$\beta=\frac{\texttt{pos\_num}+\texttt{neg\_num}}{\texttt{pos\_num}}$
corrects for the positive/negative set-size imbalance.
Stage~\uppercase\expandafter{\romannumeral1} optimizes only $\mathcal{H}$ with $\bm{\theta}$ frozen for $N_{\mathrm{sft}}$
steps (set to $5$ by default), after which the warmed-up latent state
$\mathcal{H}_{\mathrm{sft}}^{*}$ is passed to
Stage~\uppercase\expandafter{\romannumeral2}.
\vspace{-1em}
\subsection{Stage II: Latent-to-Answer Reinforcement}
\label{subsubsec:stage2}
\vspace{-0.6em}
While Stage~\uppercase\expandafter{\romannumeral1} improves the semantic
quality of $\mathcal{H}$, semantic quality alone does not guarantee that
$\mathcal{H}$ is effectively utilized during answer prediction.
Stage~\uppercase\expandafter{\romannumeral2} therefore targets latent
utilization directly, optimizing $\mathcal{H}$ to ensure that each
$\bm{h}_k$ actively contributes to the final answer.
To this end, we formulate Stage~\uppercase\expandafter{\romannumeral2} as
a reward-driven optimization over $\mathcal{H}$.
The reward is designed to enforce \textit{reasoning progression}: the
output distribution of each $\bm{h}_k$ should become increasingly
concentrated from $\bm{h}_1$ to $\bm{h}_K$, reflecting the intuition
that latent tokens grow more certain as reasoning advances along
$\mathcal{H}$.
Formally, the reward over a perturbed candidate
$\tilde{\mathcal{H}} = \mathcal{H} + \bm{\epsilon}$ is defined as:
\begin{equation}
    R(\tilde{\mathcal{H}})
    = \frac{1}{K-1}\sum_{k=1}^{K-1}
      \max\!\left(0,\; \mathcal{E}^{(k)} - \mathcal{E}^{(k+1)}\right),
    \label{eq:reward}
\end{equation}
where $\mathcal{E}^{(k)} = -\sum_{j=1}^{\delta}\tilde{p}_j^{(k)}\log\tilde{p}_j^{(k)}$ denotes the top-$\delta$ entropy at the $k$-th latent position, and $\tilde{p}_j^{(k)},\, j=1,\dots,\delta$ are the top-$\delta$ predicted token probabilities.\footnote{We set $\delta=10$ by default in all experiments.}
This reward increases when the output distribution of each $\bm{h}_k$
becomes progressively more concentrated from $\bm{h}_1$ to $\bm{h}_K$.

Optimization starts from $\mathcal{H}_{\mathrm{sft}}^{*}$ and proceeds
by stochastic local exploration for $N_{\mathrm{rl}}$ steps.
At each step $i$, a Gaussian perturbation
$\bm{\epsilon}_i \sim \mathcal{N}(\bm{0}, \sigma_i^2 \bm{I})$
is sampled to form a candidate
$\tilde{\mathcal{H}}_i = \mathcal{H}_i + \bm{\epsilon}_i$,
and the latent state is updated according to the NES gradient estimator:
\begin{equation}
    \mathcal{H}_{i+1}
    = \mathcal{H}_i
    + \frac{\alpha}{\sigma_i^2} \cdot R(\tilde{\mathcal{H}}_i) \cdot \bm{\epsilon}_i,
    \label{eq:nes_update}
\end{equation}
where $\alpha$ is the step size and $\sigma_i$ decays geometrically to
gradually shrink the exploration radius around $\mathcal{H}_i$.
Since the reward trajectory is generally non-monotonic, the
best-performing latent state encountered during optimization is retained
as $\mathcal{H}^{*}$ for final answer decoding.

\begin{algorithm}[t]
\caption{Optimization Procedure for Unsilencing Visual Latents}
\label{alg:two_stage}
\begin{algorithmic}[1]
\REQUIRE Frozen backbone $\Phi_{\bm{\theta}}$, multimodal input $(\mathcal{V},\mathcal{Q})$, initial latent state $\mathcal{H}$, warm-up steps $N_{\mathrm{sft}}$, reinforcement steps $N_{\mathrm{rl}}$
\ENSURE Optimized latent state $\mathcal{H}^{*}$
\STATE \textbf{Stage \uppercase\expandafter{\romannumeral1}: Visual Latent Warm-up}
\FOR{$n=1$ \TO $N_{\mathrm{sft}}$}
    \STATE Compute relevance scores $s_n$ via Eq.~(\ref{eq:relevance_score}) 
    \STATE Construct $\{(\mathcal{P}_k,\mathcal{N}_k)\}_{k=1}^{K}$ via Eq.~(\ref{eq:chunk_assignment}) \COMMENT{chunk-wise pos/neg assignment}
    \STATE Update $\mathcal{H}$ by minimizing Eq.~(\ref{eq:stage1_loss}) 
\ENDFOR
\STATE $\mathcal{H}_{\mathrm{sft}}^{*} \leftarrow \mathcal{H}$ 
\STATE \textbf{Stage \uppercase\expandafter{\romannumeral2}: Latent-to-Answer Reinforcement}
\STATE Initialize $\mathcal{H}_{N_{\mathrm{sft}}} \leftarrow \mathcal{H}_{\mathrm{sft}}^{*}$,\quad $\mathcal{H}^{*} \leftarrow \mathcal{H}_{N_{\mathrm{sft}}}$ \COMMENT{initialize from Stage \uppercase\expandafter{\romannumeral1} output}
\FOR{$i=N_{\mathrm{sft}}$ \TO $N_{\mathrm{sft}}+N_{\mathrm{rl}}-1$}
    \STATE Sample $\bm{\epsilon}_i \sim \mathcal{N}(\bm{0}, \sigma_i^2\bm{I})$ 
    \STATE Form candidate $\tilde{\mathcal{H}}_i \leftarrow \mathcal{H}_i + \bm{\epsilon}_i$
    \STATE Evaluate reward $R(\tilde{\mathcal{H}}_i)$ via Eq.~(\ref{eq:reward}) \COMMENT{reasoning progression reward}
    \STATE Update $\mathcal{H}_{i+1}$ via Eq.~(\ref{eq:nes_update}) 
    \IF{$R(\tilde{\mathcal{H}}_i) > R(\mathcal{H}^{*})$}
        \STATE $\mathcal{H}^{*} \leftarrow \mathcal{H}_{i+1}$ \COMMENT{retain best-performing latent state}
    \ENDIF
\ENDFOR
\RETURN $\mathcal{H}^{*}$
\end{algorithmic}
\end{algorithm}
\vspace{-0.8em}
\section{Experiments}\label{sec:experiments}
\vspace{-0.8em}
\noindent{\textbf{Implementation Details.}} All experiments are conducted with frozen MLLM backbones, and only the visual latents are optimized at inference time. Unless otherwise specified, we use Qwen2.5-VL-7B as the default backbone and set the latent reasoning length to $K=4$. In Stage I, we compute query-guided relevance scores over visual tokens and construct chunk-wise positive and negative patch sets for each latent token. We set the number of positive and negative patches to $\texttt{pos\_num}=2$ and $\texttt{neg\_num}=4$, respectively. The latent warm-up is optimized for $N_{\mathrm{sft}}=5$ steps using the contrastive latent--visual alignment objective. In Stage II, we initialize from the warmed-up latent state $\mathcal{H}_{\mathrm{sft}}^{*}$ and perform reward-driven latent optimization for $N_{\mathrm{rl}}=15$ steps. The reward encourages the predicted token distributions along the latent reasoning sequence to become progressively more concentrated, and the best latent state is retained as $\mathcal{H}^{*}$ for final answer decoding. The main experiments are run on 2 NVIDIA H100 GPUs, while the efficiency measurements and the additional experiments in \appref{app:additional} are run on 2 NVIDIA H200 GPUs. All reported results are evaluated by GPT-4o. The detailed evaluation prompt is provided in \appref{appendix:evaluation_prompt}. \\
\noindent{\textbf{Benchmarks \& Baselines.}} We evaluate across eight benchmarks spanning four categories: (1) \textit{Visual perception}: Counting, IQ Test, Relative Reflectance (RR)~\cite{fu2024blink}; (2) \textit{Visual reasoning}: MMVP~\cite{tong2024mmvp}, Hull-Bench~\cite{guan2023hallusionbench}; (3) \textit{Comprehensive}: ScienceQA~\cite{lu2022scienceqa}, MM-Star~\cite{chen2024mmstar}; (4) \textit{Mathematical reasoning}: MM-Vista~\cite{lu2024mathvista}.
We compare against two categories of baselines: \textit{non-latent reasoning} methods (MCoT~\cite{zhang2024multimodalchainofthoughtreasoninglanguage}, CCoT~\cite{Mitra_2024_CVPR}, ICoT~\cite{gao2025interleavedmodalchainofthought}) and \textit{latent reasoning} methods that operate in continuous latent space (DMLR~\cite{liu2025reasoningwithinmind}, Monet~\cite{wang2025monet}, CoVT~\cite{qin2025chain}, LVR~\cite{li2025lvr}). \new{LVR denotes the model after its supervised stage, and LVR$_{\mathrm{RF}}$ the model after its additional reinforcement-learning (RF) stage.} We use Qwen2.5-VL-3B/7B~\cite{bai2025qwen25vltechnicalreport}, R1-OneVision-7B~\cite{yang2025r1onevisionadvancinggeneralizedmultimodal}, and VLAA-Thinking-7B~\cite{chen2025sftrlearlyinvestigation} as MLLM backbones. Details of each benchmark and baseline method are provided in \appref{appendix:benchmark} and \appref{appendix:baseline}.

\vspace{-0.8em}
\subsection{Main Results}
\vspace{-0.8em}

\begin{table*}[t]
\centering
\caption{\small Comparison of different reasoning methods across multiple benchmarks. Results are reported in accuracy (\%), with bracketed values showing absolute gains over the vanilla baseline. Best and second-best results are highlighted with \colorbox{bestcell}{green} and lighter \colorbox{secondcell}{green} shading, respectively.}
\label{tab:baseline}
\vspace{-0.6em}
\setlength{\tabcolsep}{3pt}
\renewcommand{\arraystretch}{1.2}
{
\resizebox{0.99\linewidth}{!}{%
\begin{tabular}{lc *{7}{c}}
\toprule
\textbf{Methods} & 
{Counting} &
{IQTest} &
{RR} &
{MMVP} &
{Hull-Bench} &
{ScienceQA} &
{MM-Star} &
{MM-Vista} \\
\midrule
\rowcolor{gray!10}\multicolumn{9}{c}{\textit{Qwen2.5VL-7B}}\\

\textcolor{gray}{Vanilla} 
& \vanilla{65.00}
& \vanilla{22.67}
& \vanilla{38.90}
& \vanilla{68.67}
& \vanilla{65.40}
& \vanilla{82.30}
& \vanilla{59.30}
& \vanilla{58.70} \\

MCoT~\cite{zhang2024multimodalchainofthoughtreasoninglanguage}
& \res{66.67}{+1.67}
& \res{22.67}{+0.00}
& \res{35.82}{-3.08}
& \res{68.00}{-0.67}
& \res{63.62}{-1.78}
& \secondres{83.90}{+1.60}
& \res{57.90}{-1.40}
& \res{56.40}{-2.30} \\

CCoT~\cite{Mitra_2024_CVPR}
& \res{66.67}{+1.67}
& \res{21.33}{-1.34}
& \secondres{43.28}{+4.38}
& \res{69.00}{+0.33}
& \res{64.88}{-0.52}
& \res{83.80}{+1.50}
& \res{58.70}{-0.60}
& \res{57.80}{-0.90} \\

ICoT~\cite{gao2025interleavedmodalchainofthought}
& \bestres{68.33}{+3.33}
& \res{25.33}{+2.66}
& \res{37.31}{-1.59}
& \res{69.33}{+0.66}
& \res{65.51}{+0.11}
& \res{78.40}{-3.90}
& \res{60.40}{+1.10}
& \res{58.90}{+0.20} \\

DMLR~\cite{liu2025reasoningwithinmind}
& \secondres{67.50}{+2.50}
& \res{25.33}{+2.66}
& \res{34.33}{-4.57}
& \res{70.00}{+1.33}
& \res{65.83}{+0.43}
& \res{83.40}{+1.10}
& \res{60.10}{+0.80}
& \res{59.10}{+0.40} \\

Monet~\cite{wang2025monet}
& \res{60.83}{-4.17}
& \res{20.67}{-2.00}
& \res{40.36}{+1.46}
& \res{65.67}{-3.00}
& \res{65.67}{+0.27}
& \res{78.80}{-3.50}
& \res{55.50}{-3.80}
& \res{62.40}{+3.70} \\

CoVT~\cite{qin2025chain}
& \secondres{67.50}{+2.50}
& \secondres{30.67}{+8.00}
& \res{38.81}{-0.09}
& \secondres{72.67}{+4.00}
& \res{64.46}{-0.94}
& \res{83.80}{+1.50}
& \bestres{62.00}{+2.70}
& \bestres{64.30}{+5.60} \\

LVR~\cite{li2025lvr}
& \bestres{68.33}{+3.33}
& \res{28.67}{+6.00}
& \res{40.30}{+1.40}
& \res{69.00}{+0.33}
& \secondres{66.67}{+1.27}
& \res{82.80}{+0.50}
& \res{60.40}{+1.10}
& \res{61.40}{+2.70} \\

LVR$_{\mathrm{RF}}$~\cite{li2025lvr}
& \bestres{68.33}{+3.33}
& \res{27.33}{+4.66}
& \res{42.54}{+3.64}
& \res{69.67}{+1.00}
& \res{65.19}{-0.21}
& \res{83.10}{+0.80}
& \secondres{61.80}{+2.50}
& \res{62.60}{+3.90} \\

\textbf{\textit{Ours}}
& \obestres{68.33}{+3.33}
& \obestres{31.33}{+8.66}
& \obestres{44.78}{+5.88}
& \obestres{73.00}{+4.33}
& \obestres{67.30}{+1.90}
& \obestres{84.20}{+1.90}
& \osecondres{61.80}{+2.50}
& \osecondres{63.70}{+5.00} \\

\midrule
\rowcolor{gray!10}\multicolumn{9}{c}{\textit{Qwen2.5VL-3B}}\\

\textcolor{gray}{Vanilla}
& \vanilla{55.83}
& \vanilla{24.67}
& \vanilla{32.84}
& \vanilla{55.67}
& \vanilla{64.14}
& \vanilla{73.50}
& \vanilla{50.20}
& \vanilla{48.20} \\

MCoT~\cite{zhang2024multimodalchainofthoughtreasoninglanguage}
& \res{55.83}{+0.00}
& \res{21.33}{-3.34}
& \res{36.57}{+3.73}
& \res{54.33}{-1.34}
& \res{63.80}{-0.34}
& \secondres{74.10}{+0.60}
& \res{48.50}{-1.70}
& \res{47.30}{-0.90} \\

CCoT~\cite{Mitra_2024_CVPR}
& \res{54.17}{-1.66}
& \secondres{31.33}{+6.66}
& \res{27.61}{-5.23}
& \res{55.33}{-0.34}
& \res{64.03}{-0.11}
& \res{73.90}{+0.40}
& \res{49.30}{-0.90}
& \res{48.00}{-0.20} \\

ICoT~\cite{gao2025interleavedmodalchainofthought}
& \res{55.83}{+0.00}
& \res{21.33}{-3.34}
& \res{35.82}{+2.98}
& \res{56.00}{+0.33}
& \bestres{64.67}{+0.53}
& \res{68.90}{-4.60}
& \res{49.60}{-0.60}
& \res{49.80}{+1.60} \\

DMLR~\cite{liu2025reasoningwithinmind}
& \res{52.50}{-3.33}
& \res{22.67}{-2.00}
& \res{33.58}{+0.74}
& \res{56.67}{+1.00}
& \bestres{64.67}{+0.53}
& \res{73.80}{+0.30}
& \res{51.20}{+1.00}
& \bestres{51.00}{+2.80} \\

LVR~\cite{li2025lvr}
& \secondres{59.17}{+3.34}
& \res{26.67}{+2.00}
& \res{35.82}{+2.98}
& \secondres{58.67}{+3.00}
& \res{59.31}{-4.83}
& \res{73.20}{-0.30}
& \res{51.40}{+1.20}
& \res{47.50}{-0.70} \\

LVR$_{\mathrm{RF}}$~\cite{li2025lvr}
& \bestres{60.83}{+5.00}
& \res{28.33}{+3.66}
& \secondres{37.31}{+4.47}
& \res{58.33}{+2.66}
& \res{60.99}{-3.15}
& \res{73.70}{+0.20}
& \secondres{51.50}{+1.30}
& \res{48.60}{+0.40} \\

\textbf{\textit{Ours}}
& \osecondres{59.17}{+3.34}
& \obestres{32.00}{+7.33}
& \obestres{38.06}{+5.22}
& \obestres{59.33}{+3.66}
& \osecondres{64.35}{+0.21}
& \obestres{74.30}{+0.80}
& \obestres{52.40}{+2.20}
& \osecondres{50.00}{+1.80} \\
\bottomrule
\end{tabular}%
}
}
\vspace{-1.5em}
\end{table*}
\begin{table}[t]
\centering
\caption{\small Ablation study of the two-stage optimization design on Qwen2.5VL-7B.}
\setlength{\tabcolsep}{3pt}
\renewcommand{\arraystretch}{0.85}
\small
\resizebox{\linewidth}{!}{%
\begin{tabular}{ccccccc>{\revcolor}c>{\revcolor}c}
\toprule
 \textit{Stage I} & \textit{Stage II} & MMVP& Hull-Bench & ScienceQA &MM-Star &MM-Vista & MMMU & MMMU-Pro \\
\midrule
 $\bigcirc$ & $\bigcirc$ &\textcolor{gray}{68.67}& \textcolor{gray}{65.40}&\textcolor{gray}{82.30}&\textcolor{gray}{59.30}&\textcolor{gray}{58.70}&52.54&37.51\\
\checkmark &$\bigcirc$ &72.00 \textcolor{mygreen}{[+ 3.33]}&66.25 \textcolor{mygreen}{[+ 0.85]}&83.40 \textcolor{mygreen}{[+ 1.10]} &62.40 \textcolor{mygreen}{[+ 3.10]}&62.10 \textcolor{mygreen}{[+ 3.40]}&53.01 [+ 0.47]&37.05 [$-$ 0.46]\\
  \checkmark   & \checkmark& 73.00 \textcolor{deepblue}{[+ 4.33]}& 67.30 \textcolor{deepblue}{[+ 1.90]}& 84.20 \textcolor{deepblue}{[+ 1.90]} &61.80 \textcolor{deepblue}{[+ 2.50]}&63.70 \textcolor{deepblue}{[+ 5.00]}&54.31 [+ 1.77]&38.15 [+ 0.64]\\
\bottomrule
\end{tabular}}
\label{table:ablation_sft_rf}
\vspace{-1em}
\end{table}
\begin{figure}
    \centering
    \includegraphics[width=1\linewidth]{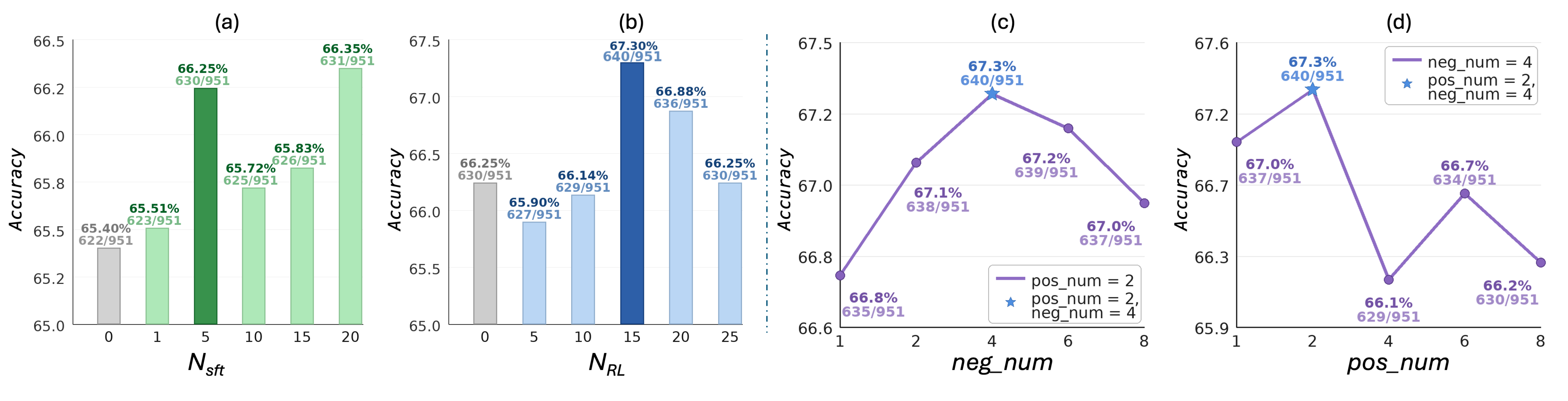}
\vspace{-2em}
\caption{\small
Ablation studies on Hull-Bench. 
(a,b) Effects of the Stage-I and Stage-II optimization steps, $N_{\mathrm{sft}}$ and $N_{\mathrm{rl}}$, respectively. 
Gray bars denote the baseline without the corresponding component, and darker bars denote the final setting; we use $N_{\mathrm{sft}}=5$ and $N_{\mathrm{rl}}=15$ for the accuracy--efficiency trade-off. 
(c,d) Effects of the negative and positive patch numbers, with \texttt{pos\_num}=2 and \texttt{neg\_num}=4 fixed, respectively. 
The results favor compact positive sets with moderate negative supervision for Stage-I latent warm-up.
}
\vspace{-1em}
    \label{fig:ablation}
\end{figure}
\autoref{tab:baseline} compares our method with representative multimodal reasoning baselines across multiple benchmarks. Overall, our method achieves the best or second-best performance on most benchmarks under both Qwen2.5VL-7B and Qwen2.5VL-3B, demonstrating the effectiveness and scalability of the proposed two-stage latent optimization framework.
On Qwen2.5VL-7B, our method achieves the best results on six out of eight benchmarks, including Counting, IQTest, RR, MMVP, Hull-Bench, and ScienceQA, while obtaining competitive second-best performance on MM-Star and MM-Vista. Compared with the vanilla baseline, it brings consistent gains across all benchmarks, with particularly notable improvements on IQTest (+8.66), RR (+5.88), MMVP (+4.33), and MM-Vista (+5.00). Compared with non-latent reasoning methods (MCoT~\cite{zhang2024multimodalchainofthoughtreasoninglanguage}, CCoT~\cite{Mitra_2024_CVPR}, ICoT~\cite{gao2025interleavedmodalchainofthought}), our method achieves consistent improvements across all benchmarks, whereas verbal chain-of-thought methods occasionally degrade below the vanilla baseline on fine-grained visual tasks, suggesting that textual reasoning alone is insufficient for these challenges. Against latent reasoning baselines, our method also shows clear advantages on perception-intensive benchmarks such as RR, MMVP, and Hull-Bench. Unlike DMLR~\cite{liu2025reasoningwithinmind}, which relies on dynamically retrieved visual patches at inference time, or training-based methods (Monet~\cite{wang2025monet}, CoVT~\cite{qin2025chain}, LVR~\cite{li2025lvr}) prone to visual input shortcuts, our method optimizes latent representations directly without parameter updates or visual re-injection, ensuring the latents serve as genuine reasoning substrates.
On Qwen2.5VL-3B, similar trends hold despite the smaller backbone capacity, with our method achieving best or second-best results on all eight benchmarks, demonstrating generalizability across model scales. \new{These gains are statistically robust: across three random seeds the standard deviation is at most 0.69 on every benchmark, and the advantage persists under sampling-based decoding (\appref{app:significance}). On the expert-knowledge benchmarks MMMU~\cite{yue2024mmmu} and MMMU-Pro~\cite{yue2024mmmupro}, our method also achieves the best accuracy (+1.77 and +0.64), while all other latent reasoning methods fall below Vanilla on MMMU-Pro (\appref{app:mmmu}).}
\vspace{-0.5em}
\subsection{Ablation Study}
\vspace{-0.5em}
\noindent\textbf{Two-Stage Optimization Design.}
We analyze the contribution of each stage and the allocation of optimization steps. As shown in \autoref{table:ablation_sft_rf}, Stage I improves MMVP from 68.67 to 72.00 and Hull-Bench from 65.40 to 66.25, indicating the benefit of enhancing latent semantics. Adding Stage II further raises performance to 73.00 and 67.30, respectively, confirming that Stage I and Stage II are largely complementary: the former improves latent quality, while the latter promotes answer-time utilization. \new{On the expert-knowledge benchmarks MMMU and MMMU-Pro, Stage II adds +1.30 and +1.10 on top of Stage I; on MMMU-Pro, it is Stage II that turns the overall gain positive.}
We also vary the optimization steps in both stages. As shown in \autoref{fig:ablation} \textcolor{deepbrick}{(a)} and \textcolor{deepbrick}{(b)}, on Hull-Bench, increasing $N_{\mathrm{sft}}$ brings moderate gains, while performance is more sensitive to $N_{\mathrm{rl}}$, peaking at $N_{\mathrm{rl}}=15$ before declining, suggesting that over-optimization in Stage II may hurt the quality of latent-guided decoding. We therefore set $N_{\mathrm{sft}}=5$ and $N_{\mathrm{rl}}=15$ for a balance between effectiveness and efficiency. \new{Stage II is also robust to its NES hyperparameters: over a broad range of perturbation scales $\sigma\in[0.05,0.2]$ and step sizes $\alpha\in[0.0005,0.002]$, the reward rises nearly monotonically and accuracy varies only marginally across seeds (\appref{app:nes}).}
\begin{figure}[h]
    \vspace{-0.7em}
    \centering
    \begin{minipage}[t]{0.55\linewidth}
        \vspace{0pt}
        \centering
        \captionsetup{type=table}
        \caption{\small Ablation on latent reasoning length on MMVP and Hull-Bench. Our method benefits more from longer latent reasoning, while LVR$_{\mathrm{RF}}$~\cite{li2025lvr} shows weaker utilization.}
        \label{tab:latent_length}
        \vspace{-0.4em}
        \footnotesize
        \setlength{\tabcolsep}{1.9pt}
        \begin{tabular}{ccccc}
        \toprule
        \multirow{2}{*}{\makecell{\footnotesize\textit{Latents} \\ \footnotesize\textit{Length $K$}}}
        & \multicolumn{2}{c}{MMVP}
        & \multicolumn{2}{c}{Hull-Bench} \\
        \cmidrule(lr){2-3} \cmidrule(lr){4-5}
        & Ours & LVR$_{\mathrm{RF}}$~\cite{li2025lvr} 
        & Ours & LVR$_{\mathrm{RF}}$~\cite{li2025lvr} \\
        \midrule
        0  & \multicolumn{2}{c}{\textcolor{gray}{68.67}}  
           & \multicolumn{2}{c}{\textcolor{gray}{65.40}} \\

        2  & 72.33 \textbf{{\scriptsize\color{deepblue!70} [+3.66]}}
           & 70.33 {\scriptsize\color{gray!70} [+1.66]}
           & 66.67 \textbf{{\scriptsize\color{deepblue!45} [+1.27]}}
           & 64.46 {\scriptsize\color{gray!55} [-0.94]} \\

        4  & 73.00 \textbf{{\scriptsize\color{deepblue!85} [+4.33]}}
           & 71.33 {\scriptsize\color{gray!90} [+2.66]}
           & 67.30 \textbf{{\scriptsize\color{deepblue!55} [+1.90]}}
           & 65.19 {\scriptsize\color{gray!82} [-0.21]} \\

        6  & 71.67 \textbf{{\scriptsize\color{deepblue!60} [+3.00]}}
           & 71.67 {\scriptsize\color{gray!100} [+3.00]}
           & 68.71 \textbf{{\scriptsize\color{deepblue!85} [+3.31]}}
           & 65.19 {\scriptsize\color{gray!82} [-0.21]} \\

        8  & 73.33 \textbf{{\scriptsize\color{deepblue!95} [+4.66]}}
           & 71.67 {\scriptsize\color{gray!100} [+3.00]}
           & 68.98 \textbf{{\scriptsize\color{deepblue!95} [+3.58]}}
           & 65.40 {\scriptsize\color{gray!90} [+0.00]} \\

        10 & 73.67 \textbf{{\scriptsize\color{deepblue!100} [+5.00]}}
           & 71.67 {\scriptsize\color{gray!100} [+3.00]}
           & 69.09 \textbf{{\scriptsize\color{deepblue!100} [+3.69]}}
           & 65.51 {\scriptsize\color{gray!95} [+0.11]} \\
        \bottomrule
        \end{tabular}
    \end{minipage}
    \hfill
    \begin{minipage}[t]{0.44\linewidth}
        \vspace{0pt}
        \centering
        \includegraphics[width=0.8\linewidth]{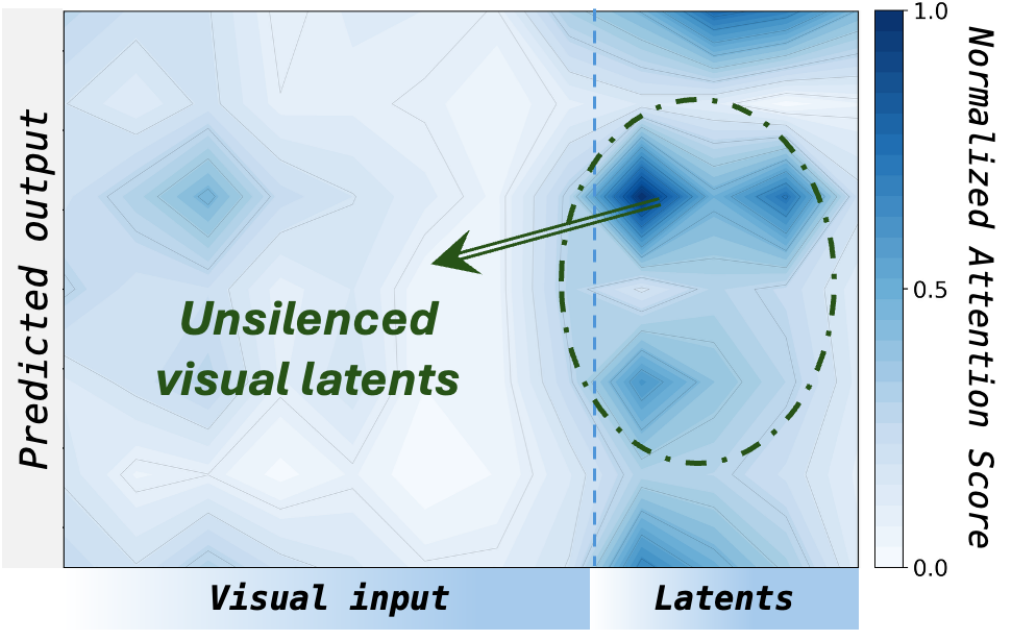}
        \captionsetup{type=figure}
    \vspace{-0.6em}
        \caption{\small Attention visualization on an MMVP sample. Optimized latents receive more focused attention than visual inputs, reducing shortcuts compared with \autoref{fig:attention_logits}.}
        \label{fig:atten_our}
    \end{minipage}
    \vspace{-1.0em}
\end{figure}

\noindent\textbf{Length of Latent Reasoning.}
We study the effect of latent reasoning length $K$ and compare with LVR$_{\mathrm{RF}}$~\cite{li2025lvr}. As shown in \autoref{tab:latent_length}, our method yields roughly consistent performance gains as $K$ increases, with MMVP improving from 72.33 to 73.67 and Hull-Bench from 66.67 to 69.09 as $K$ grows from 2 to 10, demonstrating that our method can effectively leverage additional latent reasoning capacity. In contrast, LVR$_{\mathrm{RF}}$~\cite{li2025lvr} shows little sensitivity to $K$, saturating early on MMVP and improving only marginally on Hull-Bench, which further evidences the underutilization of latent reasoning in LVR. More results on additional datasets are provided in \appref{app:more_ablation_length}.

\noindent\textbf{Contrastive Patch Configuration.}
We analyze the effect of positive and negative patch allocation in Stage I. As shown in \autoref{fig:ablation} \textcolor{deepbrick}{(c)}, performance increases steadily from 66.8\% to 67.3\% as $\texttt{neg\_num}$ grows from 1 to 4, then slightly declines, indicating that too few negatives provide insufficient contrastive signal while too many may introduce noise. As shown in \textcolor{deepbrick}{(d)}, $\texttt{pos\_num}$ has a more pronounced effect: accuracy peaks at $\texttt{pos\_num}=2$ (67.3\%) and drops notably at $\texttt{pos\_num}=4$ (66.1\%), suggesting that enlarging the positive set dilutes the alignment signal with less relevant patches. We therefore set $\texttt{pos\_num}=2$ and $\texttt{neg\_num}=4$.
\begin{table*}[t]
\centering
\caption{\small Comparison of different reasoning methods on VLAA Thinking-7B and R1 OneVision-7B. Results are reported in accuracy (\%), with bracketed values showing absolute gains over the vanilla baseline. Best and second-best results are highlighted with \colorbox{bestcell}{green} and lighter \colorbox{secondcell}{green} shading, respectively.}
\vspace{-0.6em}
\label{tab:vlaa_onevision}
\setlength{\tabcolsep}{3pt}
\renewcommand{\arraystretch}{1.0}
{
\resizebox{0.99\linewidth}{!}{%
\begin{tabular}{l *{8}{c}}
\toprule
\textbf{Methods} & 
{Counting} &
{IQTest} &
{RR} &
{MMVP} &
{Hull-Bench} &
{ScienceQA} &
{MM-Star} &
{MM-Vista} \\
\midrule

\rowcolor{gray!9}\multicolumn{9}{c}{\textit{VLAA Thinking-7B}}\\

\textcolor{gray}{Vanilla}
& \vanilla{65.00}
& \vanilla{26.67}
& \vanilla{32.84}
& \vanilla{68.30}
& \vanilla{62.00}
& \vanilla{84.50}
& \vanilla{58.90}
& \vanilla{61.10} \\

MCoT~\cite{zhang2024multimodalchainofthoughtreasoninglanguage}
& \res{62.50}{-2.50}
& \res{22.00}{-4.67}
& \res{32.84}{+0.00}
& \res{67.20}{-1.10}
& \res{62.80}{+0.80}
& \res{84.70}{+0.20}
& \res{57.10}{-1.80}
& \res{59.60}{-1.50} \\

CCoT~\cite{Mitra_2024_CVPR}
& \bestres{68.33}{+3.33}
& \res{26.00}{-0.67}
& \res{34.33}{+1.49}
& \res{68.00}{-0.30}
& \res{64.60}{+2.60}
& \res{84.60}{+0.10}
& \res{59.00}{+0.10}
& \res{60.50}{-0.60} \\

ICoT~\cite{gao2025interleavedmodalchainofthought}
& \res{60.83}{-4.17}
& \bestres{27.33}{+0.66}
& \res{32.84}{+0.00}
& \res{68.30}{+0.00}
& \res{65.90}{+3.90}
& \res{83.80}{-0.70}
& \res{58.20}{-0.70}
& \res{61.40}{+0.30} \\

DMLR~\cite{liu2025reasoningwithinmind}
& \secondres{64.17}{-0.83}
& \res{25.00}{-1.67}
& \secondres{35.07}{+2.23}
& \secondres{69.40}{+1.10}
& \secondres{67.90}{+5.90}
& \secondres{85.00}{+0.50}
& \secondres{59.20}{+0.30}
& \secondres{62.90}{+1.80} \\

\textbf{\textit{Ours}}
& \obestres{68.33}{+3.33}
& \osecondres{26.67}{+0.00}
& \obestres{35.82}{+2.98}
& \obestres{73.67}{+5.37}
& \obestres{67.93}{+5.93}
& \obestres{85.40}{+0.90}
& \obestres{62.70}{+3.80}
& \obestres{65.70}{+4.60} \\

\midrule
\rowcolor{gray!9}\multicolumn{9}{c}{\textit{R1 OneVision-7B}}\\

\textcolor{gray}{Vanilla}
& \vanilla{59.17}
& \vanilla{22.67}
& \vanilla{29.85}
& \vanilla{67.00}
& \vanilla{62.10}
& \vanilla{85.60}
& \vanilla{52.10}
& \vanilla{51.20} \\

MCoT~\cite{zhang2024multimodalchainofthoughtreasoninglanguage}
& \res{58.33}{-0.84}
& \secondres{28.00}{+5.33}
& \res{30.60}{+0.75}
& \res{68.00}{+1.00}
& \res{62.50}{+0.40}
& \bestres{86.50}{+0.90}
& \res{51.60}{-0.50}
& \res{52.50}{+1.30} \\

CCoT~\cite{Mitra_2024_CVPR}
& \secondres{60.00}{+0.83}
& \res{21.33}{-1.34}
& \res{33.58}{+3.73}
& \res{68.90}{+1.90}
& \res{63.00}{+0.90}
& \res{86.30}{+0.70}
& \res{53.50}{+1.40}
& \res{53.40}{+2.20} \\

ICoT~\cite{gao2025interleavedmodalchainofthought}
& \res{59.17}{+0.00}
& \res{22.67}{+0.00}
& \res{29.85}{+0.00}
& \res{69.60}{+2.60}
& \res{63.80}{+1.70}
& \res{85.50}{-0.10}
& \res{54.00}{+1.90}
& \res{55.60}{+4.40} \\

DMLR~\cite{liu2025reasoningwithinmind}
& \res{54.17}{-5.00}
& \res{25.33}{+2.66}
& \secondres{35.07}{+5.22}
& \bestres{71.90}{+4.90}
& \secondres{64.10}{+2.00}
& \res{83.50}{-2.10}
& \secondres{56.20}{+4.10}
& \bestres{58.00}{+6.80} \\
\textbf{\textit{Ours}}
& \obestres{61.67}{+2.50}
& \obestres{36.67}{+14.00}
& \obestres{37.30}{+7.45}
& \osecondres{71.00}{+4.00}
& \obestres{69.51}{+7.41}
& \osecondres{86.40}{+0.80}
& \obestres{56.80}{+4.70}
& \osecondres{57.30}{+6.10} \\
\bottomrule
\end{tabular}%
}
}
\vspace{-1em}
\end{table*}
\begin{figure}[t]
\centering
\begin{minipage}[c]{0.35\linewidth}
    \centering
    \includegraphics[width=\linewidth]{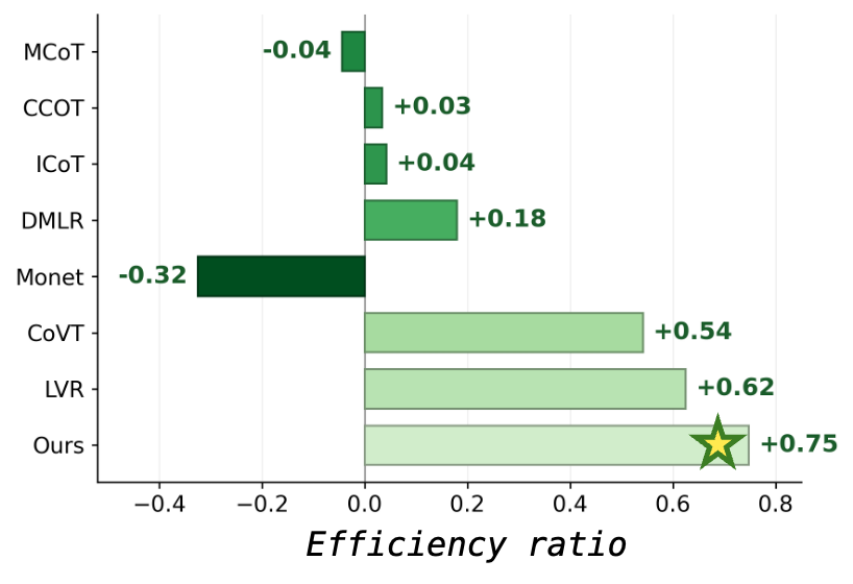}
    \vspace{-1.8em}
    \caption{\small Efficiency ratio of different methods on MMVP. Higher values indicate a better performance-efficiency balance. Calculation details are provided in \appref{app:efficiency}.}
    \label{fig:efficiency}
\end{minipage}\hfill
\begin{minipage}[c]{0.62\linewidth}
    \centering
    \revcolor
    \captionsetup{type=table}
    \caption{\small Inference overhead on MMVP (Qwen2.5-VL-7B, 2$\times$H200). TTFT: time to the first answer token, including latent optimization; WCT: end-to-end time per sample; Mem: peak GPU memory.}
    \label{tab:overhead}
    \vspace{-0.3em}
    \setlength{\tabcolsep}{3.5pt}
    \renewcommand{\arraystretch}{0.88}
    \footnotesize
    \begin{tabular}{lccccc}
    \toprule
    Method & Training-free & TTFT (s)$\downarrow$ & WCT (s)$\downarrow$ & Mem (GB)$\downarrow$ & Acc.$\uparrow$ \\
    \midrule
    Vanilla & \ding{51} & 0.24 & 2.92 & 31.18 & 68.67 \\
    MCoT & \ding{51} & 2.95 & 4.12 & 31.28 & 68.00 \\
    CCoT & \ding{51} & 3.38 & 4.56 & 31.35 & 69.00 \\
    ICoT & \ding{51} & 0.27 & 2.86 & 31.26 & 69.33 \\
    DMLR & \ding{51} & 6.67 & 7.87 & 48.09 & 70.00 \\
    Monet & \ding{55} & 0.44 & 2.30 & 31.24 & 65.67 \\
    CoVT & \ding{55} & 0.24 & 1.58 & 31.18 & 72.67 \\
    LVR & \ding{55} & 0.20 & 0.64 & 31.14 & 69.00 \\
    LVR$_{\mathrm{RF}}$ & \ding{55} & 0.20 & 0.64 & 31.14 & 69.67 \\
    \textbf{Ours} & \ding{51} & 2.56 & 3.72 & 36.10 & \textbf{73.00} \\
    \bottomrule
    \end{tabular}
\end{minipage}
\vspace{-0.6em}
\end{figure}
\begin{table}[t]
\centering
\revcolor
\caption{\small Latent-intervention sensitivity (Qwen2.5-VL-7B). Entries report $\Delta$ = Acc(intervened) $-$ Acc(original).}
\label{tab:intervention}
\setlength{\tabcolsep}{3.2pt}
\renewcommand{\arraystretch}{0.9}
\footnotesize
\begin{tabular}{l ccccc ccccc}
\toprule
\multirow{2}{*}{Method} & \multicolumn{5}{c}{Hull-Bench} & \multicolumn{5}{c}{MMVP}\\
\cmidrule(lr){2-6}\cmidrule(lr){7-11}
 & Acc. & Knockout & Noise $\sigma$=0.5 & Noise $\sigma$=1.0 & Swap & Acc. & Knockout & Noise $\sigma$=0.5 & Noise $\sigma$=1.0 & Swap\\
\midrule
LVR & 66.67 & +0.21 & +0.32 & +0.53 & +0.42 & 69.00 & +0.33 & $-$0.34 & $-$0.34 & $-$0.34\\
LVR$_{\mathrm{RF}}$ & 65.19 & +0.42 & $-$0.42 & $-$0.42 & +0.11 & 69.67 & +0.33 & $-$1.00 & $-$0.67 & 0.00\\
\textbf{Ours} & 67.30 & $-$1.90 & $-$3.36 & $-$3.47 & $-$5.26 & 73.00 & $-$4.33 & $-$5.00 & $-$5.67 & $-$7.00\\
\bottomrule
\end{tabular}
\vspace{-1em}
\end{table}
\par\noindent\textbf{Attention Visualization.}
We visualize the attention patterns on an MMVP sample in \autoref{fig:atten_our}. As expected, the optimized latents attract substantially stronger attention from the predicted outputs compared to the raw visual input tokens, and attend more to task-relevant visual regions. This confirms that the two-stage optimization effectively unsilences the visual latents, enabling them to serve as genuine reasoning substrates rather than being bypassed. More visualizations are provided in \appref{app:more_attention}.

\new{\noindent\textbf{Latent Intervention.} To verify that the unsilenced latents are causally used, we intervene on the latent tokens by removing them (Knockout), injecting Gaussian noise, or swapping them with those of a mismatched sample. As shown in \autoref{tab:intervention}, the jointly trained LVR and LVR$_{\mathrm{RF}}$ are nearly insensitive to all interventions ($|\Delta|\le 1.00$), and knockout even slightly improves their accuracy, directly confirming that their latents are silenced and their gains mainly stem from a reinforced shortcut to the raw visual input. In contrast, our accuracy drops by 1.90--5.26 on Hull-Bench and 4.33--7.00 on MMVP, showing that the optimized latents actively drive answer prediction.}

\noindent\textbf{Efficiency.}
We measure the efficiency ratio of each method as performance gain relative to output token count (details in \appref{app:efficiency}). As shown in \autoref{fig:efficiency}, our method achieves the highest ratio (+0.75), surpassing LVR (+0.62), CoVT (+0.54), and DMLR (+0.18), while non-latent methods cluster near zero and Monet yields a negative ratio ($-$0.32). This demonstrates that our method delivers superior performance gains with competitive token efficiency. \new{The comparison is fair: the optimization operates on continuous latents and generates no tokens. Since the visual input and the query are unchanged, their KV cache is computed once and reused across all optimization steps. Thus, as shown in \autoref{tab:overhead}, our method adds only about 0.8\,s per sample over Vanilla, comparable to CoT-style baselines and well below the test-time method DMLR in both time and memory, without any training (see \appref{app:overhead} for Hull-Bench).}

\noindent\textbf{Generalization to Various Models.\footnotemark}
We evaluate our two-stage latent optimization framework on VLAA Thinking-7B and R1 OneVision-7B. As shown in \autoref{tab:vlaa_onevision}, our method consistently improves over the vanilla baseline and achieves the best or second-best results on all benchmarks. Notably, it improves MMVP, Hull-Bench, and MM-Vista by +5.37, +5.93, and +4.60 on VLAA Thinking-7B, and IQTest, RR, and Hull-Bench by +14.00, +7.45, and +7.41 on R1 OneVision-7B. These results demonstrate the model-agnostic effectiveness of our framework across diverse MLLMs.
\footnotetext{Monet~\cite{wang2025monet}, CoVT~\cite{qin2025chain}, and LVR~\cite{li2025lvr} are excluded as they fundamentally alter the model's reasoning behavior via training, making them incompatible with these reasoning-capable backbones.}

\vspace{-0.6em}
\section{Conclusion}
\vspace{-0.6em}
We identify Silenced Visual Latents, a phenomenon where visual latents become semantically enhanced but are underutilized during answer prediction. To address this, we propose a frozen-backbone two-stage latent optimization framework: Stage I improves latent quality through query-guided contrastive latent--visual alignment, while Stage II enhances latent utilization via reward-driven latent-to-answer reinforcement. Experiments across multiple benchmarks and model backbones show consistent improvements, demonstrating that disentangling latent quality and latent utilization is effective for multimodal reasoning.
\vspace{-1em}
\section{Limitations}
\label{sec:limitations}
\vspace{-1em}
Our framework currently targets image-based visual reasoning tasks, and we look forward to exploring its potential in richer multimodal settings such as video understanding and embodied perception in future work. We also expect that scaling to larger and more capable backbone models will further amplify the benefits observed in our experiments.

\begin{ack}
This research is supported by Xin Zhang's A*STAR Career Development Fund (CDF) (Project No. H26-KSR0051). The authors declare no competing interests.
\end{ack}

\clearpage
{
\bibliographystyle{plain}
\bibliography{neurips_2026}
}

\clearpage
\appendix
\section{Detailed Experimental Settings}\label{app:more_ablation}
\subsection{Benchmarks}
\label{appendix:benchmark}
\noindent{\textbf{Counting} \cite{fu2024blink}.} This benchmark evaluates the MLLM abilities in detection, recognition, and compositional reasoning in complex scenes where objects may overlap, be occluded, or have varying sizes. It contains 120 test samples.

\noindent{\textbf{IQ Test} \cite{fu2024blink}.} This benchmark evaluates the ability of MLLMs to engage in graphical reasoning, without requiring any domain-specific knowledge. It contains 150 test samples.

\noindent{\textbf{Relative Reflectance} \cite{fu2024blink}.} This benchmark requires comparing the reflectance (albedo) of two pixels, where MLLMs need an understanding of material properties and their interaction with light. It contains 134 test samples.

\noindent{\textbf{MMVP}~\cite{tong2024mmvp}.} This benchmark is specially crafted to measure MLLM's visual capability via VQA, which consists of "CLIP-blind pairs" to reveal systematic visual shortcomings of MLLMs. It contains 300 test samples.

\noindent{\textbf{Hull-Bench} \cite{guan2023hallusionbench}.} This image-context reasoning benchmark evaluates MLLMs on their susceptibility to hallucination, visual illusion, and logical inconsistency through carefully designed questions. It contains 951 test samples.

\noindent{\textbf{ScienceQA} \cite{lu2022scienceqa}.} This benchmark consists of multi-modal multiple-choice questions covering diverse scientific subjects, featuring rich visual and textual contexts alongside lecture materials and explanations to assess both answer accuracy and reasoning quality. We evaluate on the first 1{,}000 samples of the test split.

\noindent{\textbf{MM-Star} \cite{chen2024mmstar}.} This vision-centric benchmark ensures visual dependency, minimal data leakage, and the requirement for advanced multi-modal capabilities. We evaluate on the first 1{,}000 samples of the test split.

\noindent{\textbf{MM-Vista} \cite{lu2024mathvista}.} This benchmark consolidates mathematical reasoning within visual contexts, encompassing diverse and complex challenges in visual perception and mathematical reasoning. It contains 1{,}000 test samples.

\new{\noindent{\textbf{MMMU}~\cite{yue2024mmmu}.} This benchmark evaluates college-level, multi-discipline multimodal understanding and reasoning that requires expert knowledge. We evaluate on the validation split, which contains 847 multiple-choice questions.}

\new{\noindent{\textbf{MMMU-Pro}~\cite{yue2024mmmupro}.} This benchmark is a more robust and difficult variant of MMMU that filters text-only answerable questions and augments the candidate options. We use the standard 10-option setting with 1{,}730 questions.}

\subsection{Reasoning Baselines}
\label{appendix:baseline}
\noindent{\textbf{MCoT}} (Multi-modal CoT)~\cite{zhang2024multimodalchainofthoughtreasoninglanguage} extends CoT reasoning to multi-modal settings by first generating a reasoning rationale conditioned on multi-modal inputs, then combining it with the original input to produce the final answer.

\noindent{\textbf{CCoT}} (Compositional CoT)~\cite{Mitra_2024_CVPR} enables multi-modal reasoning by converting the visual inputs into a JSON-formatted scene graph, based on which the reasoning is performed.

\noindent{\textbf{ICoT}} (Interleaved CoT)~\cite{gao2025interleavedmodalchainofthought} performs visual-text interleaved CoT by dynamically injecting visual clues during the reasoning process, which is triggered when specific tokens appear during reasoning.

\noindent{\textbf{LVR}}~\cite{li2025lvr} is a latent reasoning method that enables MLLMs to perform autoregressive reasoning directly in the visual embedding space. Specifically, the model is trained to generate latent states to reconstruct query-relevant visual tokens, which are interleaved with standard text generation to enhance fine-grained visual understanding. \new{LVR denotes the model after its supervised stage, and LVR$_{\mathrm{RF}}$ denotes the model after its additional reinforcement-learning (RF) stage.}

\noindent{\textbf{Monet}}~\cite{wang2025monet} trains MLLMs to perform latent visual reasoning by learning to dynamically interleave text generation with latent thinking, where the model autonomously decides when to emit a \texttt{<latent>} token to initiate visual reasoning in the embedding space.

\noindent{\textbf{CoVT}}~\cite{qin2025chain} is a training-based latent reasoning method that interleaves various continuous visual tokens into reasoning chains, including complementary perceptual cues such as segmentation, depth, and edges. Consequently, the model learns to selectively generate different types of visual tokens without requiring reinforcement learning.

\noindent{\textbf{DMLR}}~\cite{liu2025reasoningwithinmind} is a test-time latent reasoning framework that iteratively optimizes latent tokens and updates visual context through confidence-guided policy gradient optimization, dynamically injecting the most relevant visual patches into the reasoning process at inference time.

\new{\noindent{\textbf{LEAD}}~\cite{xu2026thinkinguncertainty} is a training-free latent entropy-aware decoding method that mitigates hallucinations in multimodal reasoning models by performing next-state prediction over the full token distribution carried by the hidden states. We compare with LEAD in \appref{app:lead}.}

\subsection{Evaluation Prompt}
\label{appendix:evaluation_prompt}
\begin{figure}
    \centering
    \includegraphics[width=1\linewidth]{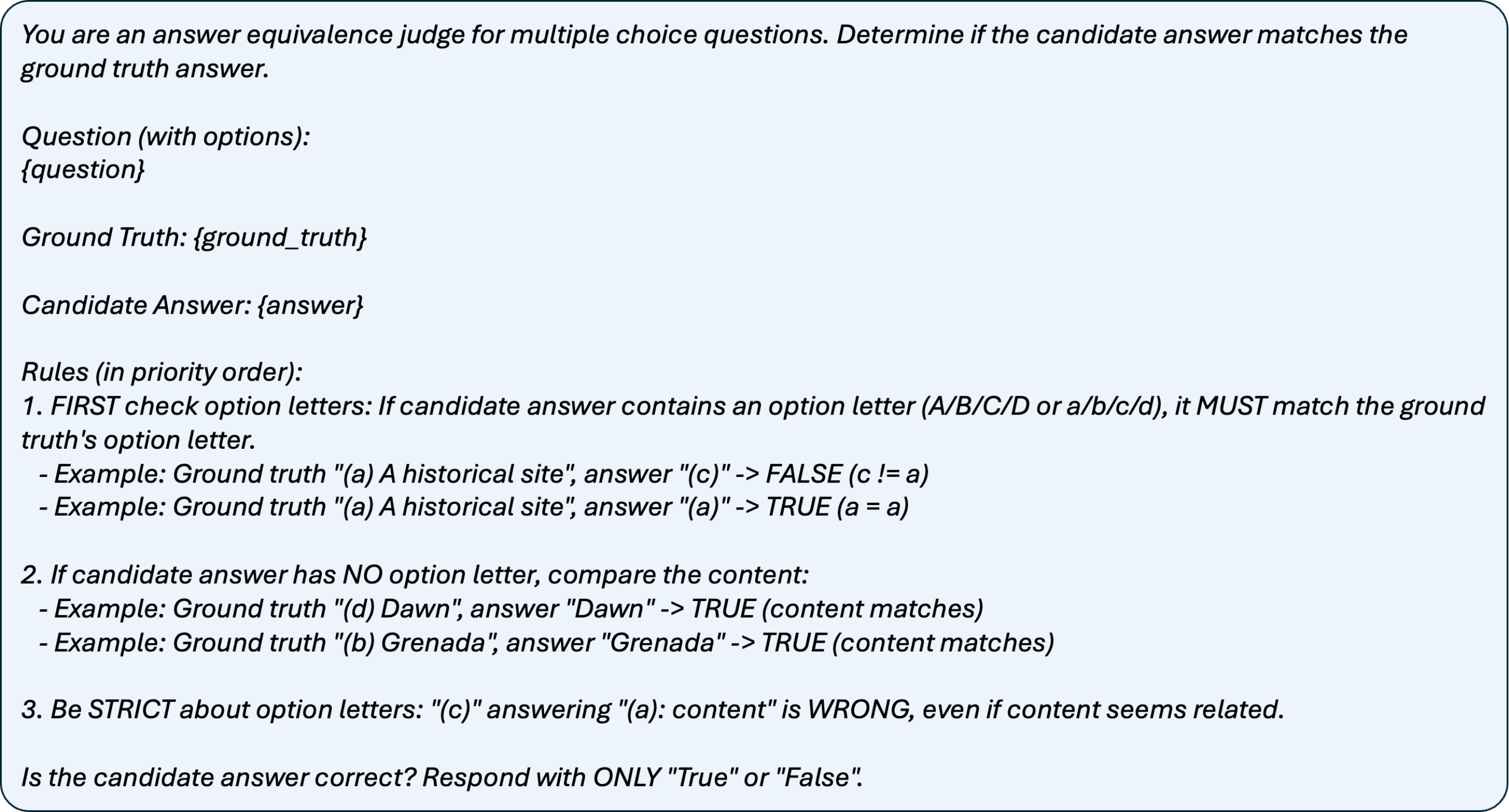}
    \caption{GPT-4o judging prompt used for answer evaluation. The prompt asks the evaluator to compare the model prediction with the ground-truth answer and judge semantic correctness.}
    \label{fig:evaluation_prompt}
\end{figure}

Following DMLR~\cite{liu2025reasoningwithinmind}, we use a fixed GPT-4o judging prompt for all GPT-based evaluations. The evaluator receives the question, the ground-truth answer, and the model prediction, and is asked to determine whether the prediction is semantically correct. This design avoids brittle string-matching and handles equivalent expressions (e.g., equivalent numerical forms or paraphrased answers) gracefully. The full prompt is shown in \autoref{fig:evaluation_prompt}.

\section{More Ablation Study}\label{app:more_ablation_study}
\subsection{Length of Latent Reasoning}\label{app:more_ablation_length}
We further evaluate the effect of latent reasoning length on ScienceQA, MM-Star, and MM-Vista. 
As shown in \autoref{tab:more_ablation_length}, our method consistently benefits from longer latent reasoning across all three benchmarks. 
For example, as the latent length increases from $K=2$ to $K=10$, performance improves from 83.40 to 85.20 on ScienceQA, from 61.30 to 62.60 on MM-Star, and from 63.00 to 64.30 on MM-Vista. 
In contrast, LVR exhibits weaker and less stable gains: it gradually drops on ScienceQA, saturates early on MM-Star, and shows only marginal variation on MM-Vista. 
These results further support that our method can more effectively exploit extended latent reasoning and translate additional latent capacity into improved answer prediction.
\begin{table}
    \centering
    \caption{Ablation on latent reasoning length on ScienceQA, MM-Star, and MM-Vista. Our method benefits more from longer latent reasoning, while LVR shows weaker utilization.}
    \small
    \setlength{\tabcolsep}{2pt}
    \begin{tabular}{ccccccc}
        \toprule
        \multirow{2}{*}{\makecell{\footnotesize\textit{Latents} \\ \footnotesize\textit{Length $K$}}}
        & \multicolumn{2}{c}{ScienceQA}
        & \multicolumn{2}{c}{MM-Star}
        & \multicolumn{2}{c}{MM-Vista}\\
        \cmidrule(lr){2-3} \cmidrule(lr){4-5} \cmidrule(lr){6-7}
        & Ours & LVR$_{\mathrm{RF}}$~\cite{li2025lvr} 
        & Ours & LVR$_{\mathrm{RF}}$~\cite{li2025lvr} 
        & Ours & LVR$_{\mathrm{RF}}$~\cite{li2025lvr}\\
        \midrule
        0  & \multicolumn{2}{c}{\textcolor{gray}{82.30}}  
           & \multicolumn{2}{c}{\textcolor{gray}{59.30}} 
           & \multicolumn{2}{c}{\textcolor{gray}{58.70}}  \\

        2  & 83.40 \textbf{{\scriptsize\color{deepblue!45} [+1.10]}}
           & 83.60 {\scriptsize\color{gray!100} [+1.30]}
           & 61.30 \textbf{{\scriptsize\color{deepblue!55} [+2.00]}}
           & 62.40 {\scriptsize\color{gray!100} [+3.10]}
           & 63.00 \textbf{{\scriptsize\color{deepblue!80} [+4.30]}}
           & 62.40 {\scriptsize\color{gray!82} [+3.70]} \\

        4  & 84.20 \textbf{{\scriptsize\color{deepblue!60} [+1.90]}}
           & 83.10 {\scriptsize\color{gray!83} [+0.80]}
           & 61.80 \textbf{{\scriptsize\color{deepblue!60} [+2.50]}}
           & 61.80 {\scriptsize\color{gray!80} [+2.50]}
           & 63.70 \textbf{{\scriptsize\color{deepblue!90} [+5.00]}}
           & 62.60 {\scriptsize\color{gray!94} [+3.90]} \\

        6  & 84.90 \textbf{{\scriptsize\color{deepblue!80} [+2.60]}}
           & 83.30 {\scriptsize\color{gray!90} [+1.00]}
           & 62.00 \textbf{{\scriptsize\color{deepblue!70} [+2.70]}}
           & 61.50 {\scriptsize\color{gray!70} [+2.20]}
           & 63.80 \textbf{{\scriptsize\color{deepblue!90} [+5.10]}}
           & 62.20 {\scriptsize\color{gray!70} [+3.50]} \\

        8  & 85.10 \textbf{{\scriptsize\color{deepblue!90} [+2.80]}}
           & 83.20 {\scriptsize\color{gray!87} [+0.90]}
           & 62.20 \textbf{{\scriptsize\color{deepblue!80} [+2.90]}}
           & 61.90 {\scriptsize\color{gray!83} [+2.60]}
           & 64.10 \textbf{{\scriptsize\color{deepblue!95} [+5.40]}}
           & 62.70 {\scriptsize\color{gray!100} [+4.00]} \\

        10 & 85.20 \textbf{{\scriptsize\color{deepblue!100} [+2.90]}}
           & 82.70 {\scriptsize\color{gray!70} [+0.40]}
           & 62.60 \textbf{{\scriptsize\color{deepblue!100} [+3.30]}}
           & 61.90 {\scriptsize\color{gray!83} [+2.60]}
           & 64.30 \textbf{{\scriptsize\color{deepblue!100} [+5.60]}}
           & 62.60 {\scriptsize\color{gray!94} [+3.90]}\\
        \bottomrule
    \end{tabular}
    \label{tab:more_ablation_length}
\end{table}
\subsection{Additional Attention Visualizations}\label{app:more_attention}
\begin{figure}[h]
    \centering
    \includegraphics[width=1\linewidth]{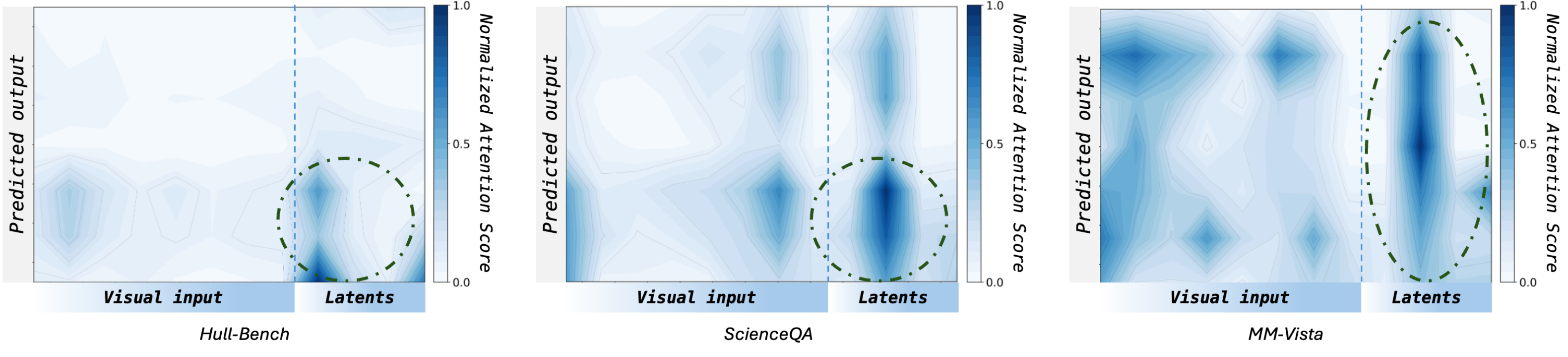}
    \caption{More attention visualizations.}
    \label{fig:more_attention}
\end{figure}

We provide more attention visualizations in \autoref{fig:more_attention} across Hull-Bench, ScienceQA, and MM-Vista. 
The optimized visual latents consistently receive concentrated attention during prediction, indicating that the model actively leverages the latent reasoning span rather than diffusely relying on raw visual inputs. 
These results further support that our two-stage optimization framework improves not only latent quality, but also their effective utilization in answer generation.
\subsection{Efficiency Ratio Calculation}\label{app:efficiency}
We define the efficiency ratio to jointly measure the performance gain and token efficiency of each method. Specifically, let $\Delta_m$ denote the average accuracy improvement of method $m$ over the vanilla baseline across all benchmarks, and let $T_m$ denote the average number of output tokens generated per sample. The efficiency ratio is defined as:
\begin{equation}
    \text{Efficiency Ratio} = \frac{\Delta_m}{T_m} \times 10
\end{equation}
where the factor of 10 is applied for readability. A higher efficiency ratio indicates that the method achieves greater performance gains per unit of output token generation. Methods with negative ratios generate more tokens while degrading performance relative to the vanilla baseline. \new{For our method, the inference-time optimization operates on continuous latent vectors and generates no tokens; answer decoding runs only once after the optimization finishes, so $T_m$ counts exactly the tokens of the final output, consistent with all other methods. The wall-clock cost of the optimization is reported separately in \appref{app:overhead}.}

\begingroup\revcolor
\section{Additional Experiments and Analyses}\label{app:additional}
Unless otherwise specified, all experiments in this section use Qwen2.5-VL-7B with the default settings in \autoref{sec:experiments}.

\subsection{Computational Overhead}\label{app:overhead}
Although our method optimizes the latents at inference time, its overhead remains moderate for three reasons. (i) The optimization is applied only to the $K$ latent tokens while the visual input and the query remain unchanged, so their prefill is performed only once and the KV cache is reused across all optimization steps; each step only recomputes the latent tokens, which is far cheaper than prefilling the full multimodal sequence. (ii) The optimization operates on continuous vectors and generates no tokens; decoding runs only once after the optimization finishes, and the high token efficiency of our method (\autoref{fig:efficiency}) yields shorter generation, offsetting much of the optimization cost. (iii) Unlike Monet, CoVT, and LVR, our method requires no training at all.
We report TTFT, end-to-end wall-clock time (WCT) per sample, and peak GPU memory on MMVP and Hull-Bench in \autoref{tab:app_overhead}, measured on the same hardware for all methods (2$\times$NVIDIA H200, fp32, 2 workers). Our method adds only about 1\,s of WCT over Vanilla on both benchmarks, comparable to CoT-style baselines and far below DMLR, the other test-time method, in both time and memory.
\begin{table}[h]
\centering
\revcolor
\caption{Computational overhead on MMVP and Hull-Bench. TTFT: time to the first token of the final answer; WCT: end-to-end time per sample; Mem: peak GPU memory; $\Delta$: absolute accuracy gain over Vanilla.}
\label{tab:app_overhead}
\setlength{\tabcolsep}{3.5pt}
\resizebox{\linewidth}{!}{%
\begin{tabular}{lc cccc cccc}
\toprule
\multirow{2}{*}{Method} & \multirow{2}{*}{Training} & \multicolumn{4}{c}{MMVP} & \multicolumn{4}{c}{Hull-Bench}\\
\cmidrule(lr){3-6}\cmidrule(lr){7-10}
 & & TTFT (s)$\downarrow$ & WCT (s)$\downarrow$ & Mem (GB)$\downarrow$ & Acc. ($\Delta$)$\uparrow$ & TTFT (s)$\downarrow$ & WCT (s)$\downarrow$ & Mem (GB)$\downarrow$ & Acc. ($\Delta$)$\uparrow$\\
\midrule
Vanilla & None & 0.24 & 2.92 & 31.18 & 68.67 (--) & 0.29 & 3.39 & 31.38 & 65.40 (--)\\
MCoT & None & 2.95 & 4.12 & 31.28 & 68.00 ($-$0.67) & 3.37 & 4.39 & 31.61 & 63.62 ($-$1.78)\\
CCoT & None & 3.38 & 4.56 & 31.35 & 69.00 (+0.33) & 4.37 & 6.15 & 32.74 & 64.88 ($-$0.52)\\
ICoT & None & 0.27 & 2.86 & 31.26 & 69.33 (+0.66) & 2.32 & 3.39 & 31.38 & 65.51 (+0.11)\\
DMLR & None & 6.67 & 7.87 & 48.09 & 70.00 (+1.33) & 8.50 & 10.16 & 55.91 & 65.83 (+0.43)\\
Monet & Required & 0.44 & 2.30 & 31.24 & 65.67 ($-$3.00) & 0.56 & 2.56 & 31.21 & 65.67 (+0.27)\\
CoVT & Required & 0.24 & 1.58 & 31.18 & 72.67 (+4.00) & 0.30 & 1.62 & 31.24 & 64.46 ($-$0.94)\\
LVR & Required & 0.20 & 0.64 & 31.14 & 69.00 (+0.33) & 0.50 & 1.64 & 31.23 & 66.67 (+1.27)\\
LVR$_{\mathrm{RF}}$ & Required & 0.20 & 0.64 & 31.14 & 69.67 (+1.00) & 0.50 & 1.64 & 31.23 & 65.19 ($-$0.21)\\
\textbf{Ours} & \textbf{None} & 2.56 & 3.72 & 36.10 & \textbf{73.00 (+4.33)} & 3.06 & 4.35 & 38.63 & \textbf{67.30 (+1.90)}\\
\bottomrule
\end{tabular}}
\end{table}

\subsection{Generalization to Expert-Knowledge Benchmarks}\label{app:mmmu}
Since Stage I aligns latents with fine-grained local visual chunks and Stage II encourages sharper predictions, a natural concern is whether the method generalizes beyond perception-grounded reasoning. We therefore evaluate on MMMU~\cite{yue2024mmmu} (validation split, 847 multiple-choice questions) and its more difficult variant MMMU-Pro~\cite{yue2024mmmupro} (standard 10-option setting, 1{,}730 questions). As shown in \autoref{tab:mmmu}, our method remains robust as the task shifts from perception toward abstract, multi-disciplinary reasoning, achieving the best accuracy on both benchmarks (+1.77 on MMMU and +0.64 on MMMU-Pro). Notably, on MMMU-Pro all other latent reasoning methods fall below Vanilla ($-$0.40 to $-$3.52), whereas ours still yields a positive gain. The stage-wise ablation on these two benchmarks is included in \autoref{table:ablation_sft_rf}.
\begin{table}[h]
\centering
\revcolor
\caption{Performance on MMMU and MMMU-Pro. $\Delta$: absolute accuracy gain over Vanilla.}
\label{tab:mmmu}
\small
\begin{tabular}{lcc}
\toprule
Method & MMMU (val) Acc. ($\Delta$)$\uparrow$ & MMMU-Pro Acc. ($\Delta$)$\uparrow$\\
\midrule
Vanilla & 52.54 & 37.51\\
MCoT & 53.96 (+1.42) & 37.86 (+0.35)\\
CCoT & 52.07 ($-$0.47) & 37.28 ($-$0.23)\\
ICoT & 44.86 ($-$7.68) & 32.14 ($-$5.37)\\
DMLR & 54.19 (+1.65) & 37.11 ($-$0.40)\\
Monet & 52.66 (+0.12) & 33.99 ($-$3.52)\\
CoVT & 50.41 ($-$2.13) & 34.16 ($-$3.35)\\
LVR & 52.30 ($-$0.24) & 34.86 ($-$2.65)\\
LVR$_{\mathrm{RF}}$ & 53.60 (+1.06) & 35.95 ($-$1.56)\\
\textbf{Ours} & \textbf{54.31 (+1.77)} & \textbf{38.15 (+0.64)}\\
\bottomrule
\end{tabular}
\end{table}

\subsection{Statistical Significance}\label{app:significance}
The only source of randomness in our method is the Gaussian perturbation in Stage II. We therefore run the full method with three random seeds under greedy decoding (\autoref{tab:seeds}): the seed-level standard deviation is at most 0.69 across all seven benchmarks, well below our gains over Vanilla. Since the main experiments use greedy decoding for reproducibility, we further evaluate under sampling-based decoding (temperature 0.7, three runs; \autoref{tab:sampling}). Sampling introduces larger run-to-run variance, yet our method retains the best mean accuracy on all four benchmarks with the smallest variance, confirming that its advantage is not an artifact of greedy decoding.
\begin{table}[h]
\centering
\revcolor
\caption{Mean $\pm$ std of our method over three random seeds (greedy decoding).}
\label{tab:seeds}
\small
\begin{tabular}{lccccccc}
\toprule
 & MMVP & Hull-Bench & ScienceQA & MM-Star & MM-Vista & MMMU & MMMU-Pro\\
\midrule
Mean & 73.11 & 67.21 & 84.30 & 61.47 & 63.43 & 53.91 & 38.17\\
Std & 0.51 & 0.51 & 0.10 & 0.31 & 0.25 & 0.69 & 0.09\\
\bottomrule
\end{tabular}
\end{table}
\begin{table}[h]
\centering
\revcolor
\caption{Comparison under sampling-based decoding (temperature 0.7, mean $\pm$ std over three runs).}
\label{tab:sampling}
\small
\begin{tabular}{lcccc}
\toprule
Method & MMVP & Hull-Bench & MMMU & MMMU-Pro\\
\midrule
Vanilla & 68.45 $\pm$ 0.69 & 65.44 $\pm$ 0.58 & 52.54 $\pm$ 0.59 & 37.47 $\pm$ 0.64\\
MCoT & 68.11 $\pm$ 1.50 & 63.72 $\pm$ 1.21 & 53.92 $\pm$ 1.13 & 37.88 $\pm$ 1.19\\
LVR & 69.67 $\pm$ 0.67 & 65.16 $\pm$ 0.58 & 53.64 $\pm$ 0.65 & 35.93 $\pm$ 0.61\\
\textbf{Ours} & \textbf{73.56 $\pm$ 0.51} & \textbf{67.33 $\pm$ 0.37} & \textbf{54.20 $\pm$ 0.35} & \textbf{38.26 $\pm$ 0.26}\\
\bottomrule
\end{tabular}
\end{table}

\subsection{Sensitivity and Stability of Stage-II Optimization}\label{app:nes}
Perturbation-based optimization in high-dimensional spaces can suffer from high variance or non-monotonic reward trajectories. To quantify this, we sweep the perturbation scale $\sigma$ and the step size $\alpha$ in \autoref{eq:nes_update} on Hull-Bench, repeating each setting with three random seeds. As shown in \autoref{tab:nes}, there is a broad safe zone ($\sigma\in[0.05,0.2]$, $\alpha\in[0.0005,0.002]$) in which the reward rises nearly monotonically and saturates within the 15-step budget, with consistently high accuracy and small seed-level variance. Across all tested settings, accuracy varies within 1.8 points and never falls below Vanilla (65.40). This is because Stage II performs only a localized refinement in the neighborhood of the Stage-I latent rather than a global search over the latent space, which largely avoids the pathologies of high-dimensional perturbation-based search. Our default setting ($\sigma=0.1$, $\alpha=0.001$) lies inside this zone.
\begin{table}[h]
\centering
\revcolor
\caption{Sensitivity to the perturbation scale $\sigma$ (with $\alpha=0.001$) and step size $\alpha$ (with $\sigma=0.1$) on Hull-Bench over three seeds. Reward: first-step $\rightarrow$ last-step reward, with the peak reward over the 15 steps in brackets.}
\label{tab:nes}
\small
\setlength{\tabcolsep}{4pt}
\begin{tabular}{lccc}
\toprule
Setting & Acc. per seed & Mean $\pm$ std & Reward (peak)\\
\midrule
$\sigma=0.01$ & 66.35 / 66.46 / 66.25 & 66.35 $\pm$ 0.11 & +8.11 $\rightarrow$ +6.53 (+12.95)\\
$\sigma=0.05$ & 67.51 / 67.61 / 67.93 & 67.68 $\pm$ 0.22 & +5.21 $\rightarrow$ +10.50 (+12.94)\\
$\sigma=0.1$ & 67.51 / 67.30 / 67.82 & 67.54 $\pm$ 0.26 & +6.98 $\rightarrow$ +12.16 (+12.16)\\
$\sigma=0.2$ & 67.93 / 67.72 / 68.03 & 67.89 $\pm$ 0.16 & +8.75 $\rightarrow$ +11.81 (+11.81)\\
$\sigma=0.5$ & 66.56 / 66.77 / 66.46 & 66.60 $\pm$ 0.16 & +11.97 $\rightarrow$ +10.92 (+11.97)\\
$\sigma=1.0$ & 66.35 / 66.04 / 66.56 & 66.32 $\pm$ 0.26 & +11.42 $\rightarrow$ +10.35 (+11.42)\\
\midrule
$\alpha=0.0001$ & 66.14 / 66.25 / 65.83 & 66.07 $\pm$ 0.22 & +7.04 $\rightarrow$ +7.20 (+7.20)\\
$\alpha=0.0005$ & 67.09 / 67.30 / 66.88 & 67.09 $\pm$ 0.21 & +7.06 $\rightarrow$ +11.20 (+11.20)\\
$\alpha=0.001$ & 67.51 / 67.30 / 67.82 & 67.54 $\pm$ 0.26 & +6.98 $\rightarrow$ +12.16 (+12.16)\\
$\alpha=0.002$ & 67.19 / 67.30 / 66.98 & 67.16 $\pm$ 0.16 & +7.18 $\rightarrow$ +10.32 (+12.86)\\
$\alpha=0.005$ & 66.35 / 66.46 / 66.56 & 66.46 $\pm$ 0.11 & +7.42 $\rightarrow$ +9.95 (+13.00)\\
$\alpha=0.01$ & 66.98 / 66.56 / 67.19 & 66.91 $\pm$ 0.32 & +7.36 $\rightarrow$ +6.50 (+12.41)\\
\bottomrule
\end{tabular}
\end{table}

\subsection{Rationale and Boundary of the Confidence-Progression Reward}\label{app:reward}
\noindent\textbf{Why confidence.} The most direct way to increase latent utilization would be to maximize the attention from the answer to the latent tokens, but this requires a full answer generation at every optimization step, which is prohibitively expensive. We instead use the prediction confidence at the latent positions, which is available before decoding. Since the backbone is frozen, any confidence gain can only come from the latents themselves, forcing answer-relevant information to be written into them, and the distribution at the last latent position directly conditions answer generation. Moreover, entropy minimization is a classical objective in test-time adaptation~\cite{wang2021tent}, confidence is known to be well calibrated with correctness in language models~\cite{kadavath2022language}, and maximizing confidence alone has been shown to improve reasoning~\cite{prabhudesai2025maximizing}. We verify the resulting link to utilization directly: after optimization, the answer attends substantially more to the latents (\autoref{fig:atten_our} and \autoref{fig:more_attention}), and corrupting the latents clearly degrades accuracy (\autoref{tab:intervention}).

\noindent\textbf{Why progression.} Entropy is not required to decrease monotonically along the latent span: the reward in \autoref{eq:reward} is zero-floored, so non-improving positions are simply not rewarded rather than penalized. Requiring confidence to increase step by step, instead of pushing every position to be uniformly sharp, preserves information diversity: each latent must add information beyond its predecessor, preventing all latents from collapsing onto the same sharp state. Consistently, a non-progressive variant that minimizes the overall entropy of the latent span shows a clear accuracy drop, although it remains above Vanilla (\autoref{tab:progression}).
\begin{table}[h]
\centering
\revcolor
\caption{Progressive vs.\ non-progressive entropy reward.}
\label{tab:progression}
\small
\begin{tabular}{lcc}
\toprule
Reward design & Hull-Bench Acc.$\uparrow$ & MMVP Acc.$\uparrow$\\
\midrule
Vanilla & 65.40 & 68.67\\
LVR & 66.67 (+1.27) & 69.00 (+0.33)\\
w/o progression (minimize overall entropy) & 66.67 (+1.27) & 71.67 (+3.00)\\
\textbf{w/ progression (\autoref{eq:reward}, ours)} & \textbf{67.30 (+1.90)} & \textbf{73.00 (+4.33)}\\
\bottomrule
\end{tabular}
\end{table}

\noindent\textbf{Failure cases.} We manually inspected the MMMU samples that the Stage-I-only variant answers correctly but the full method answers incorrectly. They are mostly calculation-type questions whose exact answer is not among the given options: the Stage-I model hedges and picks the closest option, which is often correct, whereas after Stage II the prediction becomes more decisive and commits to a specific but wrong option. In other words, the confidence prior can turn hedged-but-correct answers into confident-but-wrong ones on questions where decisiveness does not imply correctness, which marks a boundary of the confidence-progression assumption.

\subsection{Latent Initialization}\label{app:init}
The $K$ latent tokens are initialized from a special token of the vanilla MLLM. Concretely, we insert $K$ copies of the \texttt{<|endoftext|>} word embedding into the input embedding sequence immediately after the visual and query embeddings; these vectors constitute the initial latent state $\mathcal{H}_0\in\mathbb{R}^{K\times d}$, which is then optimized by the two-stage procedure. Unlike LVR, which initializes each latent from the last hidden state of the network, our latents reside in the input embedding space. To test whether $\mathcal{H}_0$ affects the optimization, we compare three initializations with all other hyperparameters fixed (\autoref{tab:init}): (A) the default \texttt{<|endoftext|>} embedding; (B) random Gaussian vectors $\mathcal{N}(0,0.02^2)$ of matched scale, which carry no semantic prior; and (C) the embedding of a different special token, \texttt{<|im\_end|>}. The three variants perform comparably. We further compare the answer-token distributions produced by latents optimized from different initializations on 50 randomly drawn MMVP samples (\autoref{tab:init_js}): the cross-initialization JS divergences (below $7\times10^{-4}$) are more than two orders of magnitude smaller than the reference divergence between different samples (0.19), with 100\% top-1 agreement versus 44\% for the reference. This indicates that, despite different starting points, the optimization converges to essentially the same solution.
\begin{table}[h]
\centering
\revcolor
\begin{minipage}[t]{0.485\linewidth}
\centering
\captionof{table}{Ablation on latent initialization.}
\label{tab:init}
\small
\setlength{\tabcolsep}{3pt}
\begin{tabular}{lcc}
\toprule
Initialization & Hull-Bench & MMVP\\
\midrule
A. \texttt{<|endoftext|>} (default) & 67.30 & 73.00\\
B. Random, matched scale & 67.82 & 72.33\\
C. \texttt{<|im\_end|>} & 66.88 & 73.00\\
\bottomrule
\end{tabular}
\end{minipage}\hfill
\begin{minipage}[t]{0.5\linewidth}
\centering
\captionof{table}{Similarity of answer-token distributions across initializations (50 MMVP samples).}
\label{tab:init_js}
\small
\setlength{\tabcolsep}{3pt}
\begin{tabular}{lcc}
\toprule
Comparison & JS (nats) & Top-1 agree.\\
\midrule
Same sample, A vs.\ B & $4.5\times10^{-4}$ & 100\%\\
Same sample, A vs.\ C & $4.6\times10^{-4}$ & 100\%\\
Same sample, B vs.\ C & $6.9\times10^{-4}$ & 100\%\\
Reference (diff.\ samples) & 0.19 & 44\%\\
\bottomrule
\end{tabular}
\end{minipage}
\end{table}

\subsection{Comparison with LEAD}\label{app:lead}
LEAD~\cite{xu2026thinkinguncertainty}, DMLR~\cite{liu2025reasoningwithinmind}, and our method share the high-level idea of exploiting the high bandwidth of latent representations to enhance multimodal reasoning without training, and both LEAD and DMLR inject visual patches to assist the final prediction. The two lines of work, however, differ in what the latents are expected to encode. LEAD inherits the latent-reasoning paradigm established for LLMs: instead of committing to the single most probable token at each decoding step, it performs next-state prediction over the full token probability distribution carried by the hidden states, retaining information that explicit text generation would discard. In contrast, DMLR, LVR, and our method instantiate latent reasoning specifically for MLLMs, expecting the latents to encode fine-grained visual evidence (a vision selector for DMLR, ROIs for LVR, and contrastive visual chunks for ours), analogous to the extension from textual CoT to multimodal CoT. The two mechanisms therefore operate at different levels and are complementary: LEAD-style entropy-aware decoding can be applied directly on top of our optimized latents.
We re-implement LEAD with its official settings and evaluate all methods under the same protocol. As shown in \autoref{tab:lead}, consistent with its hallucination-mitigation focus, LEAD shows a clear advantage on Hull-Bench, while our method delivers strong gains on both benchmarks. Combining the two (\autoref{tab:lead_combine}) yields the best Hull-Bench accuracy (69.09), and a similar gain holds for DMLR, confirming their complementarity.
\begin{table}[h]
\centering
\revcolor
\begin{minipage}[t]{0.48\linewidth}
\centering
\captionof{table}{Comparison with LEAD.}
\label{tab:lead}
\small
\setlength{\tabcolsep}{3pt}
\begin{tabular}{lcc}
\toprule
Method & MMVP ($\Delta$) & Hull-Bench ($\Delta$)\\
\midrule
Vanilla & 68.67 & 65.40\\
DMLR & 70.00 (+1.33) & 65.83 (+0.43)\\
LEAD & 69.00 (+0.33) & \textbf{67.93 (+2.53)}\\
\textbf{Ours} & \textbf{73.00 (+4.33)} & 67.30 (+1.90)\\
\bottomrule
\end{tabular}
\end{minipage}\hfill
\begin{minipage}[t]{0.48\linewidth}
\centering
\captionof{table}{Combining LEAD-style decoding with optimized latents.}
\label{tab:lead_combine}
\small
\setlength{\tabcolsep}{3pt}
\begin{tabular}{lcc}
\toprule
Setting & MMVP & Hull-Bench\\
\midrule
DMLR & 70.00 & 65.83\\
DMLR + LEAD decoding & 70.00 & 67.19\\
Ours & \textbf{73.00} & 67.30\\
Ours + LEAD decoding & 72.67 & \textbf{69.09}\\
\bottomrule
\end{tabular}
\end{minipage}
\end{table}
\endgroup

\section{Broader Impacts}\label{app:broader_impact}
This work improves the visual reasoning capabilities of MLLMs at test time without modifying model weights. As a foundational methods paper, it has no direct path to harmful applications. More capable multimodal models could in principle be misused for surveillance or disinformation, and we encourage responsible deployment with appropriate safeguards. All experiments use publicly available benchmarks and pretrained checkpoints, with no personal data involved.

\clearpage

\end{document}